\documentclass{article}
\pdfoutput=1

\usepackage{microtype}
\usepackage{graphicx}
\usepackage{subfigure}
\usepackage{amsthm}
\usepackage{amsmath}
\usepackage{nicefrac}
\usepackage{amssymb}
\usepackage{enumitem}
\usepackage{booktabs} 
\usepackage[table]{xcolor}

\theoremstyle{definition}
\newtheorem{defi}{Definition}

\theoremstyle{plain}
\newtheorem{thm}{Theorem}
\newtheorem{lem}{Lemma}
\newtheorem{cor}{Corollary}

\theoremstyle{remark}
\newtheorem{rem}{Remark}

\usepackage{multirow}
\definecolor{sky0}{HTML}{375E97}
\definecolor{sky1}{HTML}{5276AA}
\definecolor{sky2}{HTML}{1E4987}
\definecolor{sunset0}{HTML}{FB6542}
\definecolor{sunset1}{HTML}{FF8366}
\definecolor{sunset2}{HTML}{EF3C11}
\definecolor{sunflower0}{HTML}{FFBB00}
\definecolor{sunflower1}{HTML}{FFCA39}
\definecolor{sunflower2}{HTML}{C69100}

\definecolor{grass0}{HTML}{3F681C}
\definecolor{grass1}{HTML}{578134}
\definecolor{grass2}{HTML}{264809}

\newcommand{\low}{\texttt{low}}

\newcommand{\deathtimes}{\dagger}

\usepackage{hyperref}

\usepackage{algpseudocode}

\usepackage[accepted]{icml2019}

\icmltitlerunning{Connectivity-Optimized Representation Learning via Persistent Homology}

\newcommand{\mn}[1]{{\color{black}{#1}}}

\newcommand{\R}{\mathbb{R}}

\newcommand{\Z}{\mathbb{Z}}

\newcommand{\mcL}{\mathcal{L}}

\newcommand{\mcV}{\mathcal{V}}

\newcommand\restr[2]{{
  \left.\kern-\nulldelimiterspace 
  #1 
  \vphantom{\big|} 
  \right|_{#2} 
  }}

\DeclareMathOperator*{\argmin}{\arg\!\min}

\begin{document}

\twocolumn[
\icmltitle{Connectivity-Optimized Representation Learning via Persistent Homology}



\icmlsetsymbol{equal}{*}

\begin{icmlauthorlist}
\icmlauthor{Christoph D. Hofer}{sbg}
\icmlauthor{Roland Kwitt}{sbg}
\icmlauthor{Mandar Dixit}{microsoft}
\icmlauthor{Marc Niethammer}{unc}
\end{icmlauthorlist}

\icmlaffiliation{sbg}{Department of Computer Science, University of Salzburg, Austria}
\icmlaffiliation{microsoft}{Microsoft}
\icmlaffiliation{unc}{UNC Chapel Hill}
\icmlcorrespondingauthor{Christoph D. Hofer}{\texttt{chr.dav.hofer@gmail.com}}

\icmlkeywords{Machine Learning, ICML}

\vskip 0.3in
]



\printAffiliationsAndNotice{}  

\begin{abstract}
We study the problem of learning representations with controllable
connectivity properties. This is beneficial in situations when
the imposed structure can be leveraged upstream.
In particular, we control the connectivity of an autoencoder's latent
space via a novel type of loss, operating
on information from persistent homology.
Under mild conditions, this loss is differentiable and
we present a theoretical analysis of the properties induced
by the loss.
We choose one-class learning as our upstream
task and demonstrate that the imposed structure 
enables informed parameter selection for modeling
the in-class distribution via kernel density 
estimators. Evaluated on computer vision data, 
these one-class models exhibit competitive 
performance and, in a low sample size regime, 
outperform other methods by a large margin. 
Notably, our results indicate that a single 
autoencoder, trained on auxiliary (unlabeled) data, 
yields a mapping into latent space that can be
reused across datasets for one-class learning.
\vspace{-0.4cm}
\end{abstract}

\section{Introduction}
\label{section:introduction}
Much of the success of neural networks in (supervised) learning problems,
e.g., image recognition \cite{Krizhevsky12a,He16a,Huang17a},
object detection \cite{Ren15a,Liu16a,Dai16a}, or natural language processing
\cite{Graves13a,Sutskever14a} can be attributed to their
ability to learn task-specific representations, guided by a suitable loss.

In an unsupervised setting, the notion of a \emph{good/useful} representation
is less obvious. Reconstructing inputs from a (compressed)
representation is one important criterion, highlighting
the relevance of autoencoders \cite{Rumelhart86}. Other criterions
include robustness, sparsity, or informativeness 
for tasks such as clustering or classification.

To meet these criteria, the reconstruction objective is typically
supplemented by additional regularizers or cost functions that
directly (/indirectly) impose structure on the latent space. For instance,
sparse \cite{Makhzani14a}, denoising \cite{Vincent10a}, or contractive \cite{Rifai11a}
autoencoders aim at robustness of the learned representations, either
through a penalty on the encoder parametrization, or
through training with stochastically perturbed data. Additional
cost functions guiding the mapping into latent space are used in
the context of clustering, where several works \cite{Xie16a,Yang17a,Zong18a}
have shown that it is beneficial to jointly train for reconstruction
and a clustering objective. This is a prominent example for
representation learning guided towards an upstream task. Other
incarnations of imposing structure can be found in generative
modeling, e.g., using variational autoencoders \cite{Kingma14a}. Although,
in this case, autoencoders arise as a model for
approximate variational inference in a latent variable model, the additional optimization objective
effectively controls distributional aspects of the latent representations via
the Kullback-Leibler divergence. Adversarial autoencoders \cite{Makhzani16a,Tolstikhin18a}
equally control the distribution of the latent representations, but
through adversarial training.

Overall, the success of these efforts clearly shows that imposing
structure on the latent space can be beneficial. In this
work, we focus on \emph{one-class learning} as the upstream task.
This is a challenging problem, as one needs to uncover
the underlying structure of a single class using only
samples of that class.
Autoencoders are a popular backbone model
for many approaches in this area \cite{Zhou17a,Zong18a,Sabokrou18}.
By controlling \emph{topological characteristics} of the
latent representations, connectivity in particular, we argue that
kernel-density estimators can be used as effective one-class models.
While earlier works \cite{Pokorny12a,Pokorny12b} show that
informed guidelines for bandwidth selection can be derived from
studying the topology of a space, our focus is not on \emph{passively}
analyzing topological properties, but rather on \emph{actively} controlling
them. Besides work by \cite{Chen19a} on
topologically-guided regularization of \emph{decision boundaries}
(in a supervised setting), we are not aware of any other work along the
direction of backpropagating a learning signal derived from topological analyses.

\textbf{Contributions of this paper}.
\vspace{-0.3cm}
	\begin{enumerate}
	\item A novel loss, termed \emph{connectivity loss} (\S\ref{section:connectivity_loss}),
	that operates on persistence barcodes, obtained by computing  persistent homology of mini-batches. Our specific incarnation of this loss enforces a homogeneous arrangement of the representations learned by an autoencoder.

	\item Differentiability, under mild conditions, of the connectivity loss (\S\ref{subsection:differentiability}), enabling backpropagation of the loss signal through the persistent homology computation.

	\item Theoretical analysis (\S\ref{section:theory}) on the implications of controlling	connectivity via the proposed loss. 
	This reveals sample-size dependent
	densification effects that are
	beneficial upstream, e.g., for kernel-density
	estimation.

	\item One-class learning experiments (\S\ref{section:experiments}) on large-scale vision data, 
	showing that kernel-density based one-class models can be built
	on top of representations learned by a \emph{single} autoencoder.
	These representations are transferable across datasets and, in 
	a low sample size regime, our one-class models outperform 
	recent state-of-the-art methods by a large margin.	
	\end{enumerate}

\section{Background}
\label{background}

We begin by discussing the machinery to \emph{extract} connectivity information of latent representations. All proofs for the presented
results can be found in the appendix.

Let us first revisit a standard autoencoding architecture.
Given a data space $X$, we denote by $\{x_i\}, x_i \in X$, a set of training samples.
Further, let $f: X \to Z \subset \mathbb{R}^n$ and $g: Z  \subset \mathbb{R}^n \to X$ be two (non-)linear functions, referred to as the \emph{encoder} and the
\emph{decoder}.
Typically, $f$ and $g$ are parametrized by neural networks with parameters $\theta$ and $\phi$.
Upon composition, i.e., $g_{\phi} \circ f_{\theta}$, we obtain an autoencoder.
Optimization then aims to find
\begin{equation}
(\theta^*, \phi^*) = \argmin_{(\theta,\phi)} \sum_i l\Big(x_i, g_{\phi}\big(f_{\theta}(x_i)\big)\Big)\enspace,
\end{equation}
where $l: X \times X \to \mathbb{R}$ denotes a suitable \emph{reconstruction loss}.
If $n$ is much smaller than the dimensionality of $X$, autoencoder training can be thought-of as learning a (non-linear) low-dimensional embedding of $x$, i.e., $z = f_{\theta}(x)$, referred to as its \emph{latent representation}.

Our goal is to control connectivity properties of $Z$, observed via samples.
As studying connectivity requires analyzing multiple samples jointly, we 
focus on controlling the connectivity of samples in mini-batches of
fixed size.

\textbf{Notation.} We use the following notational conventions. We let $[N]$ denote the set
$\{1,\ldots,N\}$ and $\mathcal{P}([N])$ its power set.
Further, let
$B(z, r) =  \{z' \in \mathbb{R}^n: \|z-z'\| \leq r \}$ denote the closed ball of radius $r$ around $z$. By $S$, we denote a random batch of
size $b$ of latent representations $z_i = f_\theta(x_i)$.

\subsection{Filtration/Persistent homology}
To study point clouds of latent representations, $z_i$, from a topological perspective,
consider the union of closed balls (with radius $r$) around $z_i$ w.r.t. some metric $\delta$ on $\mathbb{R}^n$, i.e.,
\begin{equation}
  S_{r} = \bigcup\limits_{i=1}^b B(z_i, r)
  \text{ with } r \geq 0\enspace.
  \label{eqn:z_eps}
\end{equation}
$S_{r}$ induces a topological (sub)-space of the metric space
$(\mathbb{R}^n, \delta)$.
The number of connected components of $S_r$ is a \emph{topological property}.
A widely-used approach to access this information, grounded in algebraic topology, is to assign a growing sequence of simplicial complexes (induced by parameter $r$).
This is referred to as a \textit{filtration} and we can study how the homology groups of these complexes evolve as $r$ increases. Specifically, we study the rank of the $0$-dimensional homology groups (capturing the number of connected components) as $r$ varies.
This extension of homology to include the notion of \emph{scale} is called persistent homology
 \cite{Edelsbrunner2010}.

\begin{figure}[t!]
\centering{
\includegraphics[page=1,width=0.48\columnwidth]{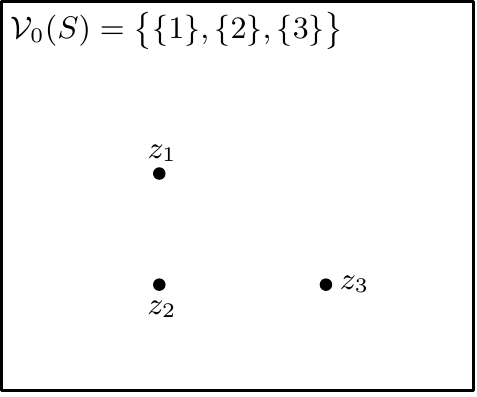}
\includegraphics[page=3,width=0.48\columnwidth]{vietoris_rips}\\  ~\includegraphics[page=4,width=0.48\columnwidth]{vietoris_rips}
\includegraphics[page=5,width=0.48\columnwidth]{vietoris_rips}}
\caption{Vietoris-Rips complex built from $S=\{z_1, z_2, z_3\}$ with only zero- and one-dimensional simplices,
i.e., vertices and edges.\label{fig:vr}}
\vspace{-0.1in}
\end{figure}

For unions of balls, the prevalent way to build a filtration is via a Vietoris-Rips complex, see Fig.~\ref{fig:vr}.
We define the Vietoris-Rips complex in a way beneficial to address differentiability and, as we only study connected components, we restrict our definition to simplices, $\sigma$, of dimension $\leq 1$. 

\vskip1.5ex
\begin{defi}[Vietoris-Rips complex]
\label{def:vr_complex}
  Let $(\mathbb{R}^n,\delta)$ be a metric space.
  For $S \subset \mathbb{R}^n$, $|S|=b$, let $\mathcal{V}(S) = \{\sigma \in \mathcal{P}([b]): 1 \leq |\sigma| \leq 2\}$ and define
  \[
  f_S: \mathcal{V}(S) \rightarrow \mathbb{R},
  \quad
  f_{S}(\sigma) =
  \begin{cases}
  0 & \sigma = \{i\}\ , \\
  \frac{1}{2}\delta(z_i, z_j) &  \sigma = \{i, j\}\enspace.
  \end{cases}
  \]
  The \emph{Vietoris-Rips complex} w.r.t. $r\geq 0$, restricted to its 1-skeleton, is defined as
  $\mathcal{V}_{r}(S) = f_S^{-1}\big((-\infty,r]\big)$.
  \end{defi}

Given that $(\varepsilon_k)_{k=1}^M$ denotes the increasing sequence of pairwise distance \emph{values}\footnote{
  Formally, $\varepsilon_k \in \{\delta(z, z'): z, z' \in S, z \neq z'\}$, $\varepsilon_k < \varepsilon_{k+1}$.
}
 of $S$ (w.r.t. $\delta$), then
\begin{equation}
  \emptyset \subset \mathcal{V}_0(S) \subset \mathcal{V}_{\nicefrac{\varepsilon_1}{2}}(S) \dots \subset \mathcal{V}_{\nicefrac{\varepsilon_M}{2}}(S)
\label{eqn:filtration}
\end{equation}
is a filtration (for convenience we set $\varepsilon_0 = 0$).
Hence, we can use $0$-dimensional persistent homology to observe the impact of $r=\nicefrac{\varepsilon}{2}$ on the connectivity of $S_{r}$, see Eq.~\eqref{eqn:z_eps}.
\vspace{-6pt}
\subsection{Persistence barcode}
Given a filtration, as in Eq.~\eqref{eqn:filtration}, $0$-dimensional persistent homology produces a multi-set of pairings $(i,j), i<j$, where  each tuple $(i,j)$ indicates a connected component that \emph{persists} from $S_{\nicefrac{\varepsilon_i}{2}}$ to $S_{\nicefrac{\varepsilon_j}{2}}$.

All $b$ points emerge in $S_0$, therefore all possible connected components appear, see Fig.~\ref{fig:vr} (top-left).
If there are two points $z_i,z_j$ contained in different connected components and $\delta(z_i,z_j) = \varepsilon_{t}$, those components \emph{merge} when transitioning from $S_{\nicefrac{\varepsilon_{t-1}}{2}}$ to $S_{\nicefrac{\varepsilon_{t}}{2}}$.
In the filtration, this is equivalent to $\mathcal{V}_{\nicefrac{\varepsilon_{t-1}}{2}}(S) \cup \{\{i,j\}\} \subset \mathcal{V}_{\nicefrac{\varepsilon_{t}}{2}}(S)$.
Hence, this specific type of connectivity information is captured by \emph{merging} events
of this form.
The $0$-dimensional \mn{\emph{persistence barcode}}, $\mathcal{B}(S)$, represents the collection of those merging events by a multi-set of tuples.
In our case, tuples are of the form $(0,\nicefrac{\varepsilon_t}{2})$, $1 \leq t \leq M$, as each tuple represents a connected component that persists from $S_0$ to $S_{\nicefrac{\varepsilon_t}{2}}$.

\vskip1.5ex
\begin{defi}[Death times]
\normalfont
\label{def:death_values}
Let $S\subset \mathbb{R}^n$ be a finite set, $(\varepsilon_k)_{k=1}^M$ be the increasing sequence of pairwise distances \emph{values} of $S$ and $\mathcal{B}(S)$ the $0$-dimensional barcode of the Vietoris-Rips filtration of $S$. We then define
\[
\deathtimes(S) = \{t: (0,\nicefrac{\varepsilon_t}{2}) \in \mathcal{B}(S) \}
\]
as the multi-set of death-times, where $t$ is contained in $\deathtimes(S)$ with the same
multiplicity as $(0,\nicefrac{\varepsilon_t}{2})$ in $\mathcal{B}(S)$.
\end{defi}

Informally, $\deathtimes(S)$ can be considered a \emph{multi-set of filtration indices}
 where merging events occur.

\vspace{-5pt}
\section{Connectivity loss}
\label{section:connectivity_loss}

To control the connectivity of a batch, $S$, of latent representations, we need (1) a suitable loss and (2) a way to compute the partial derivative of the loss with respect to its input.

Our proposed loss operates directly on $\deathtimes(S)$ with $|S|=b$.
As a thought experiment, assume that all $\varepsilon_t, t \in \deathtimes(S)$ are equal to $\eta$,
meaning that the graph defined by the 1-skeleton $\mcV_{\eta}(S)$ is connected.
For $(\varepsilon_k)_{k=1}^M$, the \emph{connectivity loss}
\begin{equation}
    \mathcal{L}_{\eta}(S) = \sum\limits_{t \in \deathtimes(S)} |\eta - \varepsilon_t|
    \label{eqn:top_reg}
\end{equation}
penalizes deviations from such a configuration. Trivially, for 
all points in $S$, there would now be at least one neighbor at 
distance $\eta$ (a beneficial property as we will see later).
The loss is optimized over mini-batches of data.
In \S\ref{section:theory}, we take into account that, in practice, $\eta$ can only be achieved \emph{approximately} and study how enforcing the proposed connectivity characteristics affects sets with cardinality larger than $b$.

\subsection{Differentiability}
\label{subsection:differentiability}

We fix $(\mathbb{R}^n, \delta) = (\mathbb{R}^n, \|\cdot\|)$,
where $\|\cdot \|$ denotes a $p$-norm and restate that $\varepsilon_t$ reflects a distance where a merging event occurs,
transitioning from $S_{\nicefrac{\varepsilon_{t-1}}{2}}$ to $S_{\nicefrac{\varepsilon_{t}}{2}}$.

In this section, we show that $\mathcal{L}_{\eta}$ is differentiable with respect to points in $S$.
This is required for end-to-end training via backpropagation, as $\varepsilon_{t}$ depends on two latent representations, $z_{i_t},z_{j_t}$, which in turn depend on the parametrization $\theta$ of $f_{\theta}$. The following definition allows us to re-formulate $\mathcal{L}_{\eta}$ to conveniently address differentiability.

\vskip1.5ex
\begin{defi}
\normalfont
\label{def:ph_indicator_function}
Let $S \subset \mathbb{R}^n$, $|S|=b$ and $z_i \in S$.
We define the indicator function
 \[
 \mathbf{1}_{i, j}(z_1, \dots, z_b) =
 \begin{cases}
 1 & \exists t \in \deathtimes(S): \varepsilon_t = || z_i - z_j ||\\
 0 & \text{else}\enspace,
 \end{cases}
 \]
 where $\{i,j\} \subset [b]$ and $(\varepsilon_k)_{k=1}^M$ is the increasing sequence of all pairwise distance \emph{values} of $S$.
\end{defi}

The following theorem states that we can compute $\mathcal{L}_{\eta}$ using Definition~\ref{def:ph_indicator_function}.
Theorem~\ref{thm:ph_indicator_loss_function_differentiable} subsequently
establishes differentiability of $\mathcal{L}_{\eta}$ using the derived reformulation.

\vskip1.5ex
\begin{thm}
\label{thm:topo_loss_with_ph_indicator_function}
Let $S \subset \mathbb{R}^n$, $|S|=b$, such that the pairwise
distances are unique. Further, let $\mathcal{L}_\eta$ be defined
as in Eq.~\eqref{eqn:top_reg} and $\mathbf{1}_{i,j}$ as
in Definition~\ref{def:ph_indicator_function}. Then,
 \[
 \mathcal{L}_{\eta}(S)
 =
 \sum\limits_{\{i, j\} \subset [b]}
 \big| \eta -  \|z_i - z_j\|\big|\cdot \mathbf{1}_{i,j}(z_1, \dots, z_b)
 \enspace.
 \]
\end{thm}

\vskip1.5ex
\begin{thm}
 \label{thm:ph_indicator_loss_function_differentiable}
 Let $S \subset \mathbb{R}^n$, $|S|=b$, such that the pairwise distances are unique.
 Then, for $1\leq u \leq b$ and $1\leq v\leq n$, the partial (sub-)derivative
 of $\mathcal{L}_\eta(S)$ w.r.t. the $v$-th coordinate of $z_u$ exists, i.e.,
 \[
  \frac{\partial \mathcal{L}_{\eta}(S)}{\partial{z_{u, v}}}
  = \sum\limits_{\{i, j\} \subset [b]}
 \frac{\partial \big| \eta -  \|z_i - z_j\|\big|}{\partial{z_{u, v}}} \cdot \mathbf{1}_{i,j}(z_1, \dots, z_b)\enspace.
 \]

\end{thm}

By using an automatic differentiation framework, such as PyTorch \cite{Paszke17a}, we can easily realize
$\mathcal{L}_{\eta}$ by implementing $\mathbf{1}_{i,j}$ from Definition~\ref{def:ph_indicator_function}.

\vskip1ex
\begin{rem}
  Theorems \ref{thm:topo_loss_with_ph_indicator_function} and \ref{thm:ph_indicator_loss_function_differentiable} require
  \emph{unique} pairwise distances, computed from $S$.
  Dropping this requirement would dramatically increase the complexity of those results, as the derivative may not be uniquely defined.
  However, under the practical assumption that the distribution of the latent representations is non-atomic, i.e.,  $P(f_{\theta}(x) = z) = 0$ for $x\in X, z \in Z$, the requirement is fulfilled almost surely.
\end{rem}

\begin{figure}
\centering{
\includegraphics[width=0.333\columnwidth]{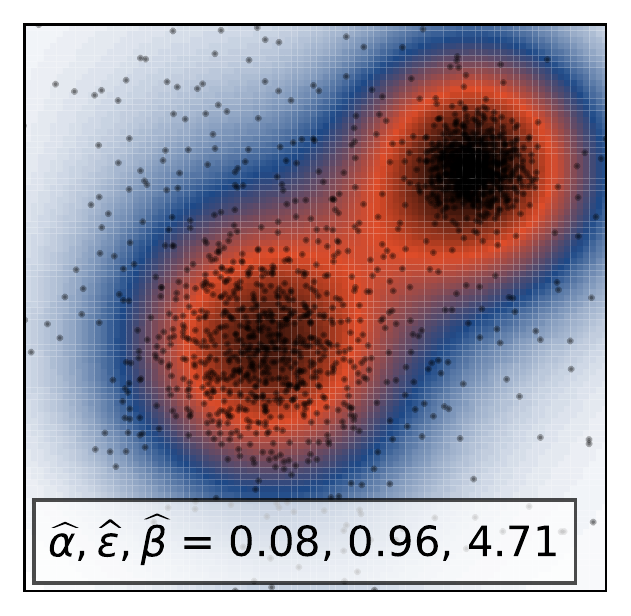}\hfill
\includegraphics[width=0.333\columnwidth]{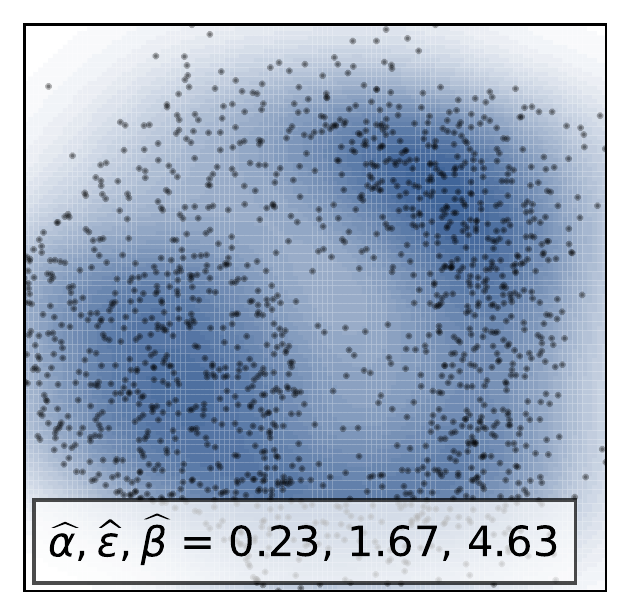}\hfill
\includegraphics[width=0.333\columnwidth]{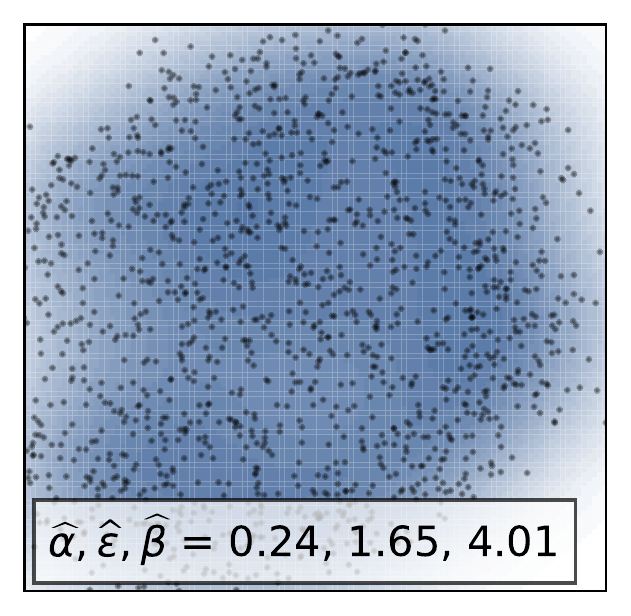}}
\vspace{-0.6cm}
\caption{\label{fig:toy}
2D toy example of a \emph{connectivity-optimized} mapping, $\texttt{mlp}: \mathbb{R}^2 \to \mathbb{R}^2$ (see \S\ref{subsection:toy}), learned on 1,500 samples, $x_i$, from three Gaussians (left). 
The figure highlights the homogenization effect enforced by the proposed loss, at 20 (middle) / 60 (right)
training epochs and lists the mean min./avg./max. values of 
$\varepsilon_t$, i.e., ($\hat{\alpha},\hat{\varepsilon},\hat{\beta}$), computed over 3,000 batches of size 50.}
%
%
%
\vspace{-0.5cm}
\end{figure}

\subsection{Toy example}
\label{subsection:toy}
We demonstrate the effect of $\mathcal{L}_{\eta}$ on toy data
generated from three Gaussians with random means/covariances, 
see Fig.~\ref{fig:toy} (left). We train a three-layer multi-layer
perceptron, $\texttt{mlp}: \mathbb{R}^2 \rightarrow \mathbb{R}^2$, 
with leaky ReLU activations and hidden layer dimensionality 20.
No reconstruction loss is used and $\mathcal{L}_{\eta}$ operates 
on the output, i.e., on fixed-size batches of $\hat{x}_i=\texttt{mlp}(x_i)$. 
Although this is different to controlling the 
latent representations, the example is sufficient to \mn{demonstrate} 
the effect of $\mathcal{L}_\eta$. The MLP is trained for 60 epochs
with batch size 50 and $\eta=2$. We then compute the mean min./avg./max. values (denoted as $\hat{\alpha}$, $\hat{\varepsilon}$, $\hat{\beta}$)
of $\varepsilon_t$ over 3,000 random batches. Fig.~\ref{fig:toy} (middle \& right)
shows the result of applying the model 
after 20 and 60 epochs, respectively.

Two observations are worth pointing out. \emph{First}, 
the gap between $\hat{\alpha}$ and $\hat{\beta}$ is fairly large, 
even at convergence. 
However, our theoretical analysis in \S\ref{section:theory} (Remark~\ref{rem:batchsizeremark}) shows that this is the expected behavior, due to the interplay between batch size and dimensionality. In this toy example, the 
range of $\varepsilon_t$ would only be small if we would train with small batch sizes (e.g., 5). 
In that case, however, gradients become increasingly unstable.
Notably, as dimensionality increases, optimizing $\mathcal{L}_\eta$  
is less difficult and effectively leads to a tighter range of 
$\varepsilon_t$ around $\eta$ (see Fig.~\ref{fig:lifetimes}). 
\emph{Second}, Fig.~\ref{fig:toy} (right) shows the desired
homogenization effect of the point arrangement, with $\hat{\varepsilon}$
close to (but smaller than) $\eta$. The latter can, to some extent, 
be explained by the previous batch size vs. dimensionality argument. 
We also conjecture that optimization is more prone to
get stuck in local minima where $\hat{\varepsilon}$ is
close to, but smaller than $\eta$. This is observed
in higher dimensions as well (cf. Fig.~\ref{fig:lifetimes}), but less prominently.

Notably, by only training with $\mathcal{L}_\eta$, we can not expect to obtain useful representations that  capture salient data characteristics as $\texttt{mlp}$ can distribute points freely, while minimizing $\mathcal{L}_{\eta}$. Hence, learning the mapping \emph{as part} of an autoencoder, optimized for reconstruction \emph{and}
$\mathcal{L}_{\eta}$, is a natural choice.

\vskip0.5ex
\emph{Intuitively, the reconstruction loss controls ``what'' is worth capturing, while the connectivity loss encourages ``how'' to topologically organize the latent representations.}

\section{Theoretical analysis}
\label{section:theory}

Assume we have minimized a reconstruction loss jointly with the connectivity loss, using mini-batches, $S$, of size $b$.
\emph{Ideally}, we obtain a parametrization of $f_\theta$ such that for every $b$-sized random sample, it holds that $\varepsilon_t$ equals $\eta$ for $t \in \deathtimes(S)$.
Due to two competing optimization objectives, however, we can only expect $\varepsilon_t$ to lie in an interval $[\alpha,\beta]$ around $\eta$.
This is captured in the following definition.

\vskip1.5ex
\begin{defi}[$\alpha$-$\beta$ connected set]
\normalfont
\label{defn:alphabeta}
  Let $S \subset \mathbb{R}^n$ be a finite set and let  $(\varepsilon_k)_{k=1}^M$ be the increasing sequence of pairwise distance \emph{values} of $S$.
  We call $S$ \emph{$\alpha$-$\beta$-connected} iff
  \[
  \alpha = \min\limits_{t \in \deathtimes(S)} \varepsilon_t \quad\text{ and }\quad
  \beta = \max \limits_{t \in \deathtimes(S)} \varepsilon_t\enspace.
  \]
\end{defi}
If $S$ is $\alpha$-$\beta$ connected, all merging events of connected components occur during the transition from $S_{\nicefrac{\alpha}{2}}$ to $S_{\nicefrac{\beta}{2}}$.

Importantly, during training, $\mathcal{L}_\eta$ \emph{only} controls properties of $b$-sized subsets \emph{explicitly}. Thus, at convergence, $f_{\theta}(S)$ with $|S|=b$ is $\alpha$-$\beta$ connected.
When building upstream models, it is desirable to understand how the latent representations are affected for samples \emph{larger} than $b$.

To address this issue, let $B(z, r)^0 =  \{z' \in \mathbb{R}^n: \|z-z'\| < r \}$ denote the interior of $B(z,r)$ and let $B(z, r, s) = B(z, s) \setminus B(z, r)^0$ with $r<s$ denote the \emph{annulus} around $z$. In the following, we formally investigate the impact of $\alpha$-$\beta$ connectedness on
the density around a latent representation.
The next lemma captures one particular \emph{densification} effect that occurs
if sets larger than $b$ are mapped via a learned $f_{\theta}$.

\vskip1.5ex
\begin{lem}
  \label{lem:onion_ring}
  Let $2\leq b \leq m$ and $M \subset \mathbb{R}^n$ with $|M| = m$ such that for each $S \subset M$ with $|S| = b$, it holds that $S$ is $\alpha$-$\beta$-connected.
  Then, for $d = m - b$ and $z \in M$ arbitrary but fixed, we find
  $M_z \subset M$ with $|M_z| = d + 1$ and $M_z \subset B(z, \alpha, \beta)$.
\end{lem}

Lemma~\ref{lem:onion_ring} yields a \emph{lower bound}, $d+1$, on the number of points in the annulus around $z \in M$. However, 
it does not provide any further insight whether there may or may not  exist more points of this kind. Nevertheless, the density around $z \in M$ increases with $|M|=m$, for $b$ fixed. 

\vskip1.5ex
\begin{defi}[$d$-$\varepsilon$-dense set]
  Let $S \subset \mathbb{R}^n$ and $\varepsilon > 0$.
  We call $S$ \emph{$\varepsilon$-dense} iff~
  $\forall z \in S~\exists z' \in S\setminus\{z\} : \|z-z'\| \leq \varepsilon$.
  For $d \in \mathbb{N}$, we call $S$ $d$-$\varepsilon$-\emph{dense} iff
  ~$\forall z \in S$
  \begin{equation*}
  \begin{gathered}
  \exists M \subset S \setminus \{z\}:
  |M|= d,\ z' \in M \Rightarrow \|z-z'\| \leq \varepsilon \enspace.\\
  \end{gathered}
  \end{equation*}
\end{defi}

The following corollary of Lemma~\ref{lem:onion_ring} provides insights into the density behavior of samples around points $z \in M$.

\vskip1.5ex
\begin{cor}
\label{cor:onionring_cor2}
  Let $2\leq b \leq m$ and $M \subset \mathbb{R}^n$ with $|M| = m$ such that for each $S \subset M$ with $|S| = b$, it holds that $S$ is $\alpha$-$\beta$-connected. Then
  $M \text{ is } (m - b + 1)\text{-}\beta\text{-dense}$.
\end{cor}

Informally, this result can be interpreted
as follows: Assume we have optimized for a specific $\eta$.
At convergence, we can collect $\varepsilon_t$ for $t \in \deathtimes(S)$ over batches (of size $b$) in the last training epoch to estimate $\alpha$ and $\beta$ according to Definition~\ref{defn:alphabeta}.
Corollary~\ref{cor:onionring_cor2} now quantifies how many neighbors, i.e., $m-b+1$, within
distance $\beta$ can be found around each $z \in M$. We exploit this insight
in our experiments to construct kernel density estimators with an informed choice
of the kernel support radius, set to the value $\eta$ we optimized for.

We can also study the implications
of Lemma~\ref{lem:onion_ring} on the \emph{separation} of points in $M$.
Intuitively, as $m$ increases, we expect the separation of points in $M$ to decrease, as densification occurs.
We formalize this by drawing a connection to the concept of \emph{metric entropy}, see \cite{Tao14a}.

\vskip1.5ex
\begin{defi}[$\varepsilon$-metric entropy]
\label{defn:metricentropy}
  Let $S \subset \mathbb{R}^n$, $\varepsilon > 0$. We call $S$
  \emph{$\varepsilon$-separated} iff~$\forall z,z' \in S: z \neq z' \Rightarrow \|z-z'\| \geq \varepsilon$.
  For $X \subset \mathbb{R}^n$, the \emph{$\varepsilon$-metric entropy} of $X$ is defined as
  \[
  N_{\varepsilon}(X) = \max\{|S|: S \subset X \text{ and } S \text{ is } \varepsilon\text{-separated}\}
  \enspace.
  \]
\end{defi}

Setting $\mathcal{E}_{\alpha,\beta}^{\varepsilon,n} = N_{\varepsilon}\big(B(0,\alpha, \beta)\big)$, i.e., the metric entropy of the annulus in $\mathbb{R}^n$, allows formulating a second corollary of Lemma~\ref{lem:onion_ring}.

\vskip1.5ex
\begin{cor}
      \label{cor:onionring_cor1}
  Let $2\leq b \leq m$ and $M \subset \mathbb{R}^n$ with $|M| = m$ such that for each $S \subset M$ with $|S| = b$, it holds that $S$ is $\alpha$-$\beta$-connected.
  Then, for $\varepsilon > 0$ and $m-b+1 > \mathcal{E}^{\varepsilon,n}_{\alpha, \beta}$,
  it follows that $M$ is not $\varepsilon$-separated.
  \end{cor}

Consequently, understanding the behavior of $\mathcal{E}^{\varepsilon,n}_{\alpha, \beta}$ is important, specifically in relation to the dimensionality, $n$, of the latent space.
To study this in detail, we have to choose a specific $p$-norm. We use $\|\cdot\|_1$ from now on, due to its better behavior in high dimensions, see \cite{Aggarwal01a}.

\vskip1.5ex
\begin{lem}
\label{lem:onion_ring_entropy}
  Let $\varepsilon < 2\alpha$ and $\alpha < \beta$. Then, in
  $(\mathbb{R}^n, \|\cdot\|_1)$, it holds that
  $\mathcal{E}^{\varepsilon,n}_{\alpha, \beta}
  \leq
  (\nicefrac{2\beta}{\varepsilon}+1)^n -
  (\nicefrac{2\alpha}{\varepsilon}-1)^n$.
\end{lem}

This reveals an exponential dependency on $n$, in other words, a manifestation of the \emph{curse of dimensionality}.
Furthermore, the bound in Lemma~\ref{lem:onion_ring_entropy} is not sharp, as it is based on a volume argument (see appendix).
Yet, in light of Corollary~\ref{cor:onionring_cor1}, it 
yields a conservative guideline to assess whether $M$ is large enough to be no longer $\varepsilon$-separated.
In particular, let $|M|=m$ and set $\varepsilon=\eta$.
If
\begin{equation}
\label{eqn:practicalbound}
m-b+1 >
(\nicefrac{2\beta}{\eta}+1)^n -
(\nicefrac{2\alpha}{\eta}-1)^n\enspace,
\end{equation}
then $M$ is not $\eta$-separated, by virtue of Lemma~\ref{lem:onion_ring_entropy}.

In comparison to the densification result of Corollary~\ref{cor:onionring_cor2}, we obtain no
quantification of separatedness for \emph{each} $z \in M$.
We can only guarantee that beyond a certain sample size, $m$, \emph{there exist} two points
with distance smaller than $\varepsilon$.

\vskip2ex
\begin{rem}
\label{rem:batchsizeremark}
We can also derive \emph{necessary} conditions on the
size $b=|S|$, given $\alpha,\beta,\eta$ and $n$, such that $M$
satisfies the conditions of Lemma~\ref{lem:onion_ring}. In particular, 
assume that the conditions are satisfied and set $|M|=m=2b-1$.
Hence, we can find $M_z$ with $M_z \subset M \cap B(z, \alpha, \beta)$ and 
$|M_z| = d+1 = m-b+1 = b=|S|$ for $z\in M$.
As every $b$-sized subset is $\alpha$-$\beta$-connected, it 
follows that $M_z$ is $\alpha$-$\beta$-connected, in particular, 
$\alpha$-separated. This yields the 
necessary condition $b \leq \mathcal{E}^{\alpha, n}_{\alpha, \beta}$. 
By applying Lemma \ref{lem:onion_ring_entropy} with $\varepsilon = \alpha$, we get
$b \leq (\nicefrac{2\beta}{\alpha}+1)^n - 1$, establishing a relation between $b, \alpha, \beta$ and $n$. 
For example, choosing $b$ large, in relation to $n$, results in an increased gap 
between $\widehat{\alpha}$ and $\widehat{\beta}$, as seen in Fig.~\ref{fig:toy} (for 
$b=50, n=2$ fixed). 
Increasing $n$ in relation to $b$ tightens this gap, 
as we will later see in \S\ref{sec:one-class-learning}.

\end{rem}


\section{Experimental study}
\label{section:experiments}

We focus on \emph{one-class learning} for visual data, i.e., building classifiers for single classes, using only data from that class.

\noindent
\textbf{Problem statement}.
\mn{Let} $C \subset X$ be a class from the space of images, $X$, from which a sample $\{x_1,\ldots,x_m\} \subset C$ is available.
Given a new sample, $y_* \in X$, the goal is to identify whether this sample belongs to $C$.
It is customary to ignore the actual binary classification task, and consider
a \emph{scoring function} $s: X\to \mathbb{R}$ instead.
Higher scores indicate membership in $C$.
We further assume access to an unlabeled \emph{auxiliary dataset}.
This is reasonable in the context of visual data, as such data is readily available.

\textbf{Architecture \& Training}.
We use a convolutional autoencoder \mn{following the DCGAN encoder/discriminator architecture of \cite{Radford16a}.} The encoder has three convolution layers (followed by Leaky ReLU \mn{activations}) with $3 \times 3$ filters, applied with a stride of $2$.
From layer to layer, the number of filters (initially, 32) is doubled.

\begin{figure}
\centering{
\includegraphics[scale=0.98]{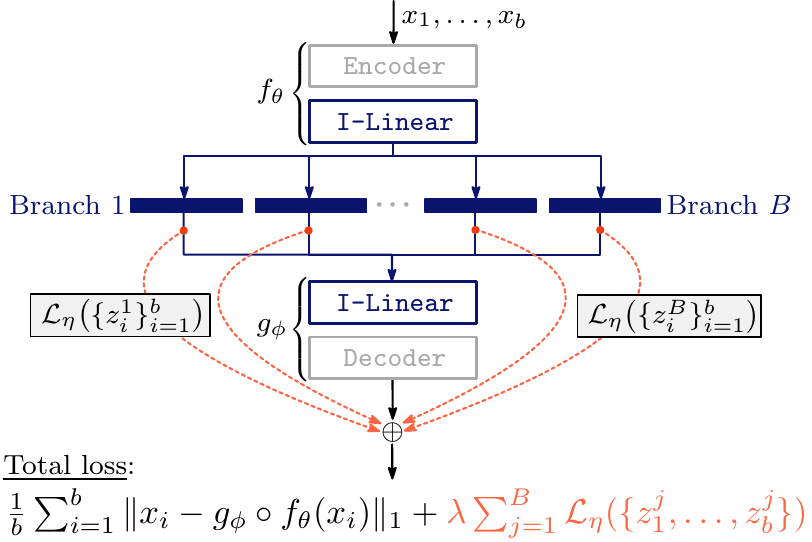}}
\vspace{-0.1cm}
\caption{
\emph{Autoencoder architecture} with $B$ independent branches mapping into latent space $Z \subset \mathbb{R}^n =  \mathbb{R}^{D} \times \cdots \times \mathbb{R}^{D}$.
The \textcolor{sunset0}{connectivity loss $\mathcal{L}_{\eta}$} is computed per branch, summed, and added to the reconstruction loss 
(here $\|\cdot\|_1$).\label{fig:arch}}
\vspace{-0.4cm}
\end{figure}

The output of the last convolution layer \mn{is mapped} into the latent space $Z \subset \mathbb{R}^n$ via a restricted variant of a linear layer (\texttt{I-Linear}).
The weight matrix $W$ of this layer \mn{is block-diagonal}, \mn{corresponding} to $B$ branches, independently mapping into $\mathbb{R}^{D}$ with $D=n/B$. 
Each  branch has its own connectivity loss, operating on the $D$-dimensional representations. This is motivated by the dilemma that we need  dimensionality (1) \emph{sufficiently high} to capture the underlying characteristics of the data and (2) \emph{low enough} to effectively optimize connectivity (see \S\ref{subsection:parameteranalysis}).
The decoder mirrors the encoder, using convolutional transpose operators \cite{Zeiler10a}.
The full architecture is shown in Fig.~\ref{fig:arch}.

For optimization, we use Adam \cite{Kingma14} with a fixed learning rate of $0.001$, $(\beta_1,\beta_2) = (0.9,0.999)$ and a batch-size of 100.
The model is trained for $50$ epochs.

\noindent
\textbf{One-class models}.
As mentioned in \S\ref{section:introduction}, our goal is to build  one-class models that leverage the structure imposed on the latent representations.
To this end, we use \mn{a simple non-parametric approach.}
Given $m$ training instances, $\{x_i\}_{i=1}^{m}$, of a new class $C$, we first compute $z_i = f_{\theta}(x_i)$ and then split $z_i$ into its $D$-dimensional parts $z_i^1,\ldots,z_i^B$, provided by each branch (see Fig.~\ref{fig:arch}).
\mn{For} a test sample $y_*$, we compute its latent representation $z_* = f_\theta(y_*)$ and its corresponding parts $z_*^1,\ldots,z_*^B$.
\mn{The \emph{one-class score} for $y_*$ is defined as}
\begin{equation}
s(y_*) = \sum_{j=1}^{B} \left|\left\{z_i^j: \|z_*^j - z_i^j\|\leq \eta, 1\leq i\leq m \right\}\right|
\enspace,
\label{eqn:scoring}
\end{equation}
where $\eta$ is the value previously used to learn $f_{\theta}$; for one 
test sample this scales with $\mathcal{O}(Bm)$.
For each branch, Eq.~\eqref{eqn:scoring} \emph{counts} how many of the stored training points of class $C$ lie in the $\|\cdot\|_1$-ball of radius $\eta$ around $z_*$.
If normalized, this constitutes a non-parametric kernel density estimate with a uniform kernel of radius $\eta$.
\emph{No optimization, or parameter tuning, is required to build such a model}.
The scoring function only \mn{uses} the imposed connectivity structure.
Given enough training samples (i.e., $m>b$), Corollary~\ref{cor:onionring_cor1} favors that the set of training points within a ball of radius $\eta$ around $z_*$ is non-empty.

\subsection{Datasets}

\textbf{CIFAR-10/100}.
CIFAR-10 \cite{Krizhevsky09a} contains 60,000 natural images of size $32\times 32$ in 10 classes. 5,000 images/class are available for training, 1,000/class for validation.
CIFAR-100 contains the same number of images, but consists of 100 classes (with little class overlap to CIFAR-10).
For comparison to other work, we also use the \emph{coarse} labels of CIFAR-100, where all 100 classes are aggregated into 20 coarse categories (\textbf{CIFAR-20}).

\textbf{Tiny-ImageNet}.
This dataset represents a medium scale image corpus of 200 visual categories with 500 images/class available for training, 50/class for validation and 50/class for testing.
For experiments, we use the training and validation portion, as labels for the test set are not available.

\textbf{ImageNet}. For large-scale testing, we 
use the ILSVRC 2012 dataset \cite{Deng09a} which consists of 1,000 classes with $\approx$1.2 million images for training ($\approx 1281$/class on avg.) and 50,000 images (50/class) for validation.

All images are resized to $32 \times 32$ (ignoring non-uniform aspect ratios) and normalized to range $[0,1]$.
We resize to $32 \times 32$ to ensure that autoencoders trained on, e.g., CIFAR-10/100, can be used for one-class experiments on ImageNet.

\vspace{-5pt}
\subsection{Evaluation protocol}

To evaluate one-class learning performance on one dataset, we only train a \emph{single} autoencoder on the
unlabeled auxiliary dataset to obtain $f_{\theta}$.
E.g., our results on Tiny-ImageNet and ImageNet use the \emph{same} autoencoder trained on CIFAR-100.
The experimental protocol follows \cite{Ruff18a} and \cite{Golan18a}.
Performance is measured via the area under the ROC curve (AUC) which is a common choice \cite{Iwata16a,Golan18a,Ruff18a}.
We use a \emph{one-vs-all} evaluation scheme.
Assume we have $N$ classes and want to evaluate one-class performance on class $j$.
Then, a one-class model is built from $m$ randomly chosen samples of class $j$. For evaluation, all test samples of class $j$ are assigned a label of $1$;
all other samples are assigned label $0$. The AUC is computed from the scores
provided by Eq.~\eqref{eqn:scoring}.
This is repeated for all $N$ classes and the AUC, averaged over (1) all classes and (2) five runs (of randomly picking $m$ points) is reported.

\vspace{-5pt}
\subsection{Parameter analysis}
\label{subsection:parameteranalysis}

We fix the dataset to CIFAR-100 and focus on the aspects of latent 
space dimensionality, the weighting of $\mathcal{L}_{\eta}$ and 
the transferability of the connectivity characteristics\footnote{We fix
$\eta=2$ throughout our experiments.}.

\begin{figure}[h!]
\centering{
\includegraphics[height=2.8cm]{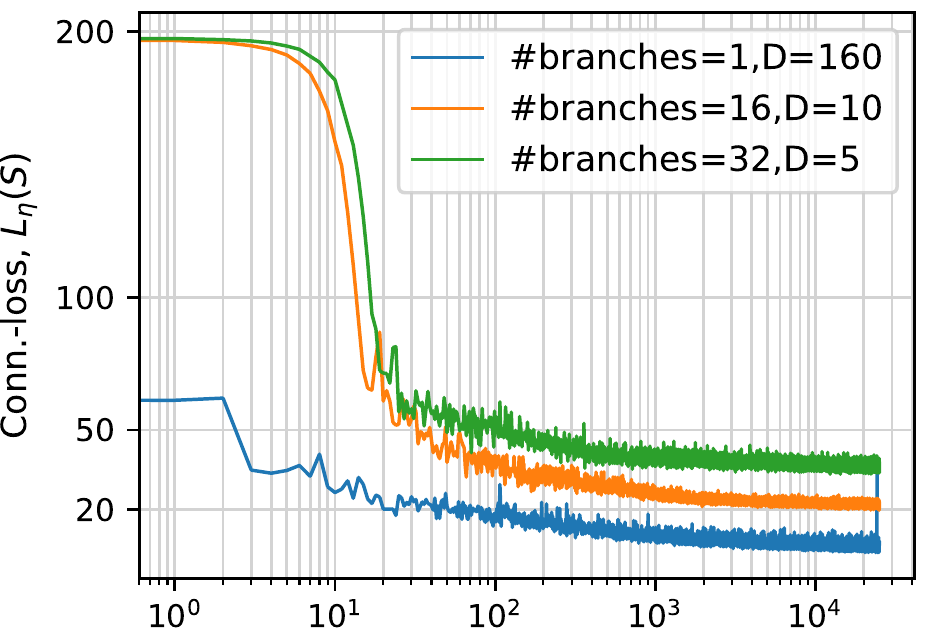}\hfill
\includegraphics[height=2.8cm]{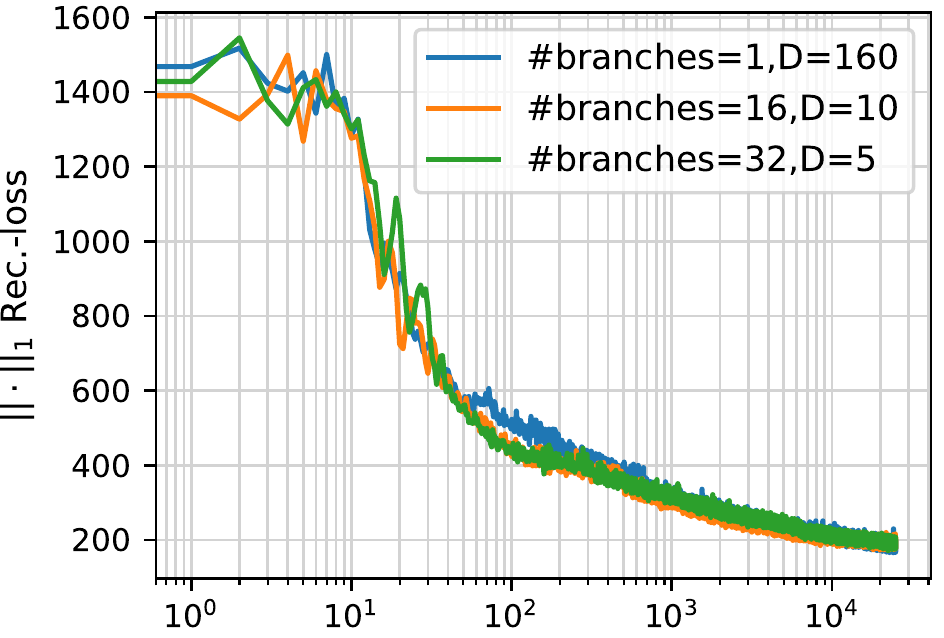}}
\vspace{-0.2in}
\caption{Connectivity (left) and reconstruction (right) loss over all 
training iterations on CIFAR-100 w/ and w/o branching.
\label{fig:trainingcurve}}
\vspace{-0.1in}
\end{figure}

\emph{First}, it is important to understand the interplay between the latent
dimensionality and the constraint imposed by $\mathcal{L}_{\eta}$.
On the one hand, a low-dimensional space allows fewer
possible latent configurations without violating the desired
connectivity structure. On the other hand, as dimensionality
increases, the concept of proximity degrades quickly for
$p$-norms \cite{Aggarwal01a}, rendering the connectivity optimization problem
trivial. Depending on the dataset, one also needs
to ensure that the underlying data characteristics 
are still captured. To balance these objectives, we divide
the latent space into sub-spaces (via separated branches). Fig.~\ref{fig:trainingcurve}
(left)
shows an example where the latent dimensionality is fixed (to $160$), but
branching configurations differ. As expected, the connectivity loss
\emph{without} branching is small, even at initialization. 
In comparison, models \emph{with} separate branches exhibit high 
connectivity loss initially, but the loss decreases rapidly throughout training.
Notably, the reconstruction error, see Fig.~\ref{fig:trainingcurve} (right),
is almost equal (at convergence) across all models.

Thus, with respect to reconstruction, the latent space carries equivalent
information with and without branching, but is \emph{structurally different}.
Further evidence is provided when using
$f_\theta$ for one-class learning
on CIFAR-10.
Branching leads to an average
AUC of 0.78 and 0.75 (for 16/32 branches), while
no branching yields an AUC of
0.70. This indicates that controlling connectivity
in low-dimensional subspaces leads to a structure
beneficial for our one-class models.

\begin{figure}[h!]
\centering{
\includegraphics[height=2.8cm]{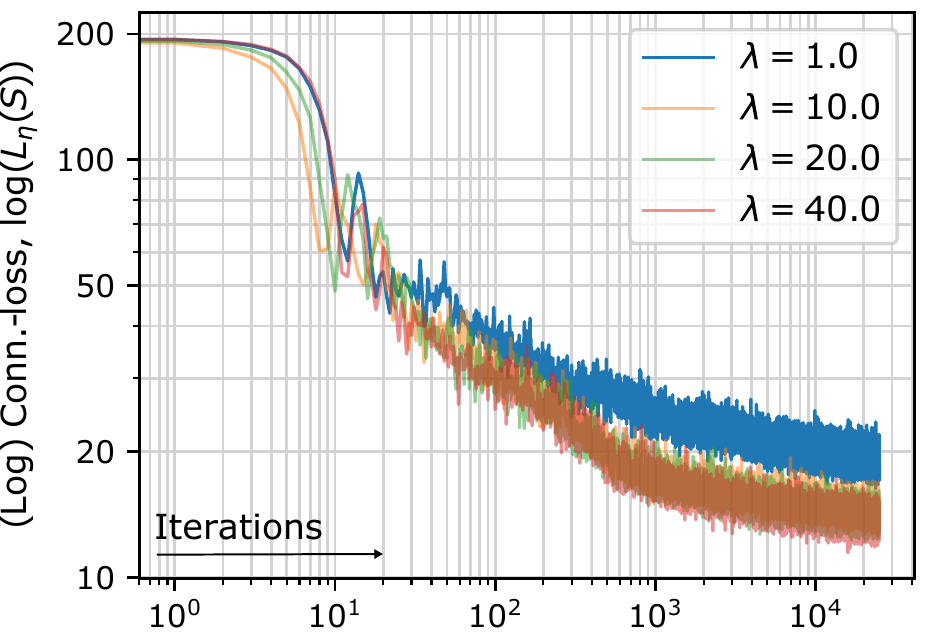}\hfill
\includegraphics[height=2.8cm]{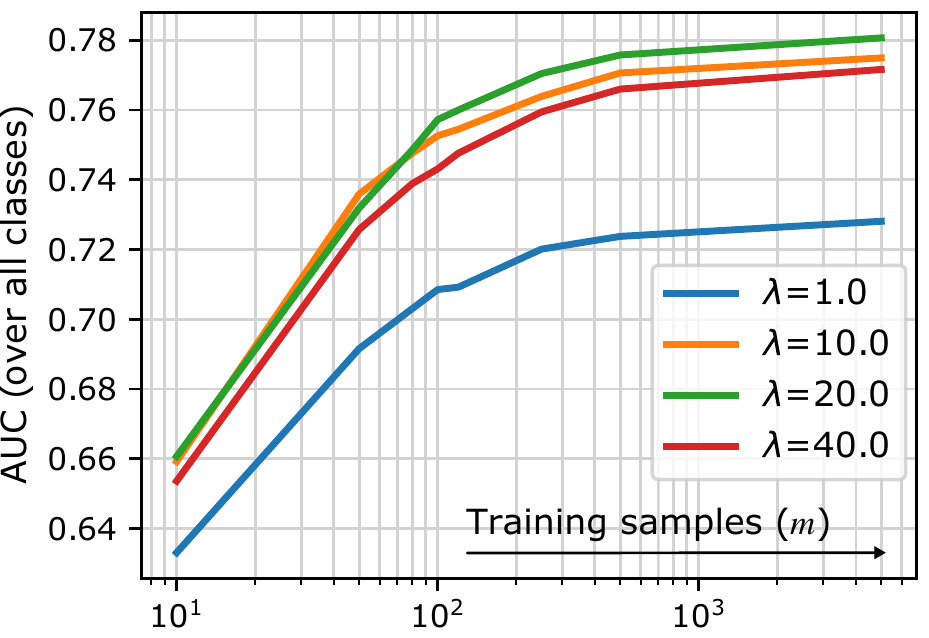}}
\vspace{-0.1in}
\caption{(Left) Connectivity loss over training
iterations on CIFAR-100 for $16$ branches and
varying $\lambda$; 
(Right) One-class performance (AUC) on CIFAR-10 over
the number of training samples, $10 \leq m \leq$ 5,000, per class.
\label{fig:weighting}}
\end{figure}

\emph{Second}, we focus on the branching architecture and
study the effect of weighting $\mathcal{L}_{\eta}$ via $\lambda$. Fig.~\ref{fig:weighting} (left) shows the connectivity loss
over all training iterations on CIFAR-100 for four different
values of $\lambda$ and $16$ branches.

\vspace{-0.2cm}

During training, the behavior of $\mathcal{L}_{\eta}$ is almost equal for $\lambda\geq 10.0$. For $\lambda=1.0$, however, the loss
noticeably converges to a higher value. In fact, reconstruction
error dominates in the latter case, leading to a less homogeneous
arrangement of latent representations. This detrimental effect
is also evident in Fig.~\ref{fig:weighting} (right) which shows the average AUC
for one-class learning on CIFAR-10 classes as a function of
the number of samples used to build the kernel density estimators.

\begin{figure}[h!]
\centering{
\includegraphics[width=0.99\columnwidth]{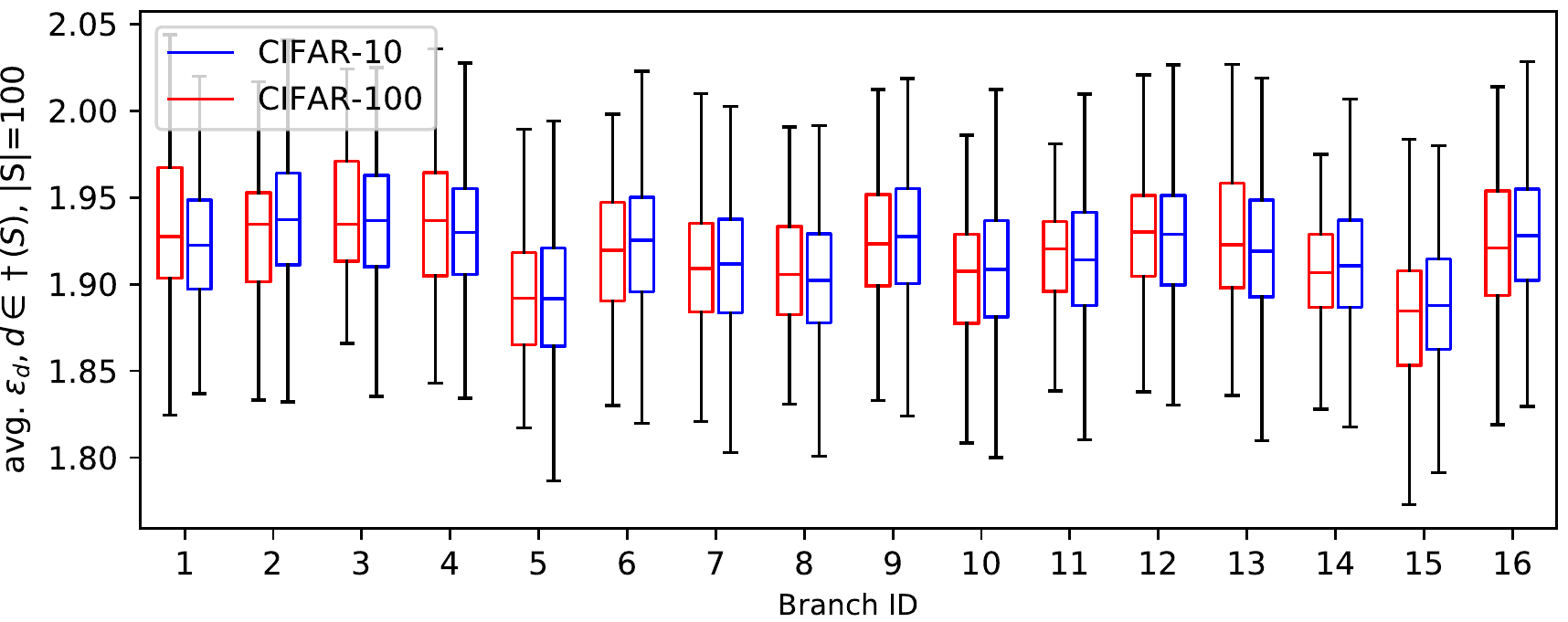}}
\vspace{-0.1in}
\caption{Average $\varepsilon_d, d \in \dagger(S)$, per branch, computed from batches, $S$, of size 100 over
CIFAR-10 (all) and CIFAR-100 (test split); $f_\theta$ is learned from the training portion of CIFAR-100.\label{fig:lifetimes}}
\vspace{-0.14in}
\end{figure}

\emph{Finally}, we assess whether the properties induced by
$f_\theta$, learned on auxiliary data (CIFAR-100), generalize to
another dataset (CIFAR-10). To this end, we
train an autoencoder with $16$ sub-branches and $\lambda=20$. We then compute the average death-times
per branch using batches of size 100 on (i) the test split of CIFAR-100
and (ii) over all samples of CIFAR-10. Fig.~\ref{fig:lifetimes}
shows that the distribution of death-times is
consistent \emph{within} and \emph{across} datasets. 
Also, the increased dimensionality (compared to our 2D toy example) per branch  leads to (i) 
a tight range of death-times  
and (ii) death-times closer to $\eta=2$,
consistent with Remark~\ref{rem:batchsizeremark}.

\subsection{One-class learning performance}
\label{sec:one-class-learning}

Various incarnations of one-class problems occur throughout the literature,
mostly in an anomaly or novelty detection context; see \cite{Pimentel14a} for a survey.
Outlier detection \cite{Xia15a,You17a} and out-of-distribution detection \cite{Hendrycks17a,Liang18a,Lee18a} are related tasks, but the
problem setup is different. The former works under the premise of corrupted
data, the latter considers a dataset as \emph{one} class.

We compare against recent state-of-the-art approaches, including \mn{techniques using autoencoders and techniques that do not.} 
In the \textbf{DSEBM} approach of \cite{Zhai16a}, the density of one-class samples
is modeled via a \mn{deep structured energy model}. 
The energy function then serves as a scoring criterion.
\textbf{DAGMM} \cite{Zong18a} follows a similar objective, but, as in our approach, density estimation
is performed in \mn{an autoencoder's latent space}. Autoencoder
and density estimator, i.e., a Gaussian mixture model (GMM), are trained jointly. The
negative log-likelihood under the GMM is then used for scoring. \mn{\textbf{Deep-SVDD} \cite{Ruff18a} is conceptually different.} \mn{Here, the idea} of support vector data description (SVDD) from \cite{Tax04a} is extended to neural
networks. An encoder (pretrained in an autoencoder setup) is
trained to map one-class samples into a hypersphere with minimal radius
and fixed center. The distance to this center is used for scoring.
Motivated by the observation that softmax-scores of trained multi-class
classifiers tend to differ between \mn{in- and out-of-distribution} samples
\cite{Hendrycks17a}, \cite{Golan18a} recently proposed a technique
(\textbf{ADT}) based on \emph{self-labeling}. In particular, a neural network classifier 
is trained
to distinguish among $72$ geometric transformations applied to one-class samples.
For scoring, each transform is applied to new samples and the softmax outputs (of the
class corresponding to the transform) of this classifier are averaged.

Non-linear dimensionality reduction via autoencoders
also facilitates \mn{using} \emph{classic} approaches to one-class problems, e.g., one-class SVMs \cite{Schoelkopf01}. We compare against such a baseline, \textbf{OC-SVM (CAE)}, using the latent representations of a convolutional autoencoder (CAE).

\begin{table}[t!]
\caption{
AUC scores for one-class learning, averaged over all
classes and 5 runs. 
ADT-$m$ and Ours-$m$ denote that only $m$ training
samples/class are used. The dataset in parentheses denotes the \emph{auxiliary} dataset on which $f_\theta$ is trained. 
All std. deviations for our method are within $10^{-3}$ and $10^{-4}$.
\label{table:overall}}
\small
\vspace{1pt}
\begin{center}
\begin{tabular}{llc}
\toprule
\textbf{Eval. data.} & \textbf{Method} & \textbf{AUC} \\
\midrule
\multirow{10}{*}{CIFAR-10}  & OC-SVM (CAE)          & $0.62$ \\
              & DAGMM \cite{Zong18a}      & $0.53$ \\
              & DSEBM \cite{Zhai16a}      & $0.61$ \\
              & Deep-SVDD \cite{Ruff18a}    & $0.65$ \\
              & ADT \cite{Golan18a}     & $\mathbf{0.85}$ \\
              & \textcolor{sky0}{\emph{Low sample-size regime}}   & \\
              \cmidrule{2-3}
              & ADT-120           & $0.69$ \\
              & ADT-500           & $0.73$ \\
              & ADT-1,000           & $0.75$ \\
              & \textbf{Ours}-120 (CIFAR-100)   & $\mathbf{0.76}$ \\
\midrule
\multirow{10}{*}{CIFAR-20}  & OC-SVM (CAE)          & $0.63$ \\
              & DAGMM \cite{Zong18a}      & $0.50$ \\
              & DSEBM \cite{Zhai16a}      & $0.59$ \\
              & Deep-SVDD \cite{Ruff18a}    & $0.60$ \\
              & ADT \cite{Golan18a}     & $\mathbf{0.77}$ \\
              & \textcolor{sky0}{\emph{Low sample-size regime}}   & \\
                            \cmidrule{2-3}
              & ADT-120             & $0.66$ \\
              & ADT-500             & $0.69$ \\
              & ADT-1,000             & $0.71$ \\
              & \textbf{Ours}-120 (CIFAR-10)    & $\mathbf{0.72}$ \\
              \midrule
\multirow{2}{*}{CIFAR-100}  & ADT-120              &  $0.75$  \\
              & \textbf{Ours}-120 (CIFAR-10) &  $\mathbf{0.79}$  \\
              \midrule
\multirow{2}{*}{Tiny-ImageNet}  & \textbf{Ours}-120 (CIFAR-10)  &  $0.73$  \\
                  & \textbf{Ours}-120 (CIFAR-100) &  			   $0.72$  \\
              \midrule
\multirow{2}{*}{ImageNet}  & \textbf{Ours}-120 (CIFAR-10) &        $0.72$  \\
                		   & \textbf{Ours}-120 (CIFAR-100)   &     $0.72$  \\
              \bottomrule
\end{tabular}
\end{center}
\vspace{-0.4cm}
\end{table}

\noindent
\textbf{Implementation.} For our approach\footnote{\scriptsize\url{https://github.com/c-hofer/COREL\_icml2019}}, we fix the latent 
dimensionality to 160 (as in \S~\ref{subsection:parameteranalysis}), 
use 16 branches and set $\lambda=20$ (the encoder, $f_\theta$, has $\approx$800k parameters). We implement a PyTorch-compatible GPU variant of the persistent homology computation, i.e., Vietoris-Rips construction and matrix reduction (see appendix). For all reference methods, except Deep-SVDD,
we use the implementation(s) provided by \cite{Golan18a}. OC-SVM (CAE) and DSEBM use a DCGAN-style
convolutional encoder with slightly more parameters ($\approx$1.4M)
than our variant and 256 latent dimensions.
DAGMM relies on the same encoder, a latent dimensionality of five and three GMM components. 

\noindent
\textbf{Results.}
Table~\ref{table:overall} lists the AUC score (averaged over classes and 5 runs) obtained on each dataset.
For our approach, the name in parentheses denotes the auxiliary (unlabeled) dataset used to learn $f_{\theta}$. 

\emph{First}, ADT exhibits the best performance on CIFAR-10/20.
However, if one aims to thoroughly assess one-class performance,
testing \mn{on} CIFAR-10/20 can be misleading, as the variation in the out-of-class
samples is limited to 9/19 categories.
\mn{Hence,} it is desirable to evaluate on datasets with higher out-of-class variability, e.g., ImageNet.
In this setting, the bottleneck of all other methods is the requirement of optimizing \emph{one model/class}.
In case of ADT, e.g., one Wide-ResNet \cite{Zagoruyko16a} with 1.4M parameters needs to be trained per class. On ImageNet, this amounts to a total of 1,400M parameters (spread
over 1,000 models). On one GPU (Nvidia GTX 1080 Ti) this requires $\approx$75 hrs.
Our approach requires to train $f_{\theta}$ only once, e.g., on CIFAR-100 and $f_{\theta}$ 
can be reused across datasets.

\emph{Second}, CIFAR-10/20 contains a large number of training samples/class. 
As the number of classes increases, training set size per class typically drops, e.g., to $\approx$1,000 on
ImageNet. We \mn{therefore} conduct a second experiment, studying the impact of training
set size per class on ADT. Our one-class models are built from a 
fixed sample size of 120, which is slightly higher than the training batch size (100), thereby 
implying densification (by our results of \S\ref{section:theory}).
We see that performance of ADT drops rapidly from 0.85 to 0.69 AUC on CIFAR-10 and from 0.77 to 0.66 on CIFAR-20 when only 120 class samples are used.
Even for 1,000 class samples, ADT performs slightly worse than our approach.
Overall, in this \emph{low sample-size regime}, our one-class models seem to clearly benefit from
the additional latent space structure.

\emph{Third}, to the best of our knowledge, we report the first full evaluation of
one-class learning on CIFAR-100, Tiny-ImageNet and ImageNet. This is possible as $f_{\theta}$ is reusable across datasets and
the one-class models do not require optimization.
For CIFAR-100, we also ran ADT with 120 samples to establish a fair comparison.
Although this requires training 100 Wide-ResNet models, it is still possible at reasonable effort.
Importantly, our method maintains performance when moving from
Tiny-ImageNet to full ImageNet, indicating beneficial scaling behavior with respect to 
the amount of out-of-class variability in a given dataset.

\vspace{-0.1in}
\section{Discussion}
\label{section:discussion}

We presented \emph{one} possibility for controlling topological / geometric properties of an autoencoder's latent space. The  
connectivity loss is tailored to enforce beneficial properties for one-class learning.
We believe this to be a key task that clearly reveals the usefulness of a representation.
Being able to backpropagate through a loss based on persistent homology has broader implications.
For example, other types of topological constraints may be useful for a wide range of tasks, such as clustering.
From a theoretical perspective, we show that controlling connectivity allows establishing \emph{provable} results for latent space densification and separation.
Composing multi-class models from one-class models (cf.~\cite{Tax08a}), built on top of a topologically-regularized representation, is another promising direction.

\clearpage

\section*{Acknowledgements}

This research was supported by NSF ECCS-1610762, the Austrian Science Fund (FWF project P 31799) and the Spinal Cord Injury and Tissue Regeneration Center Salzburg (SCI-TReCS), Paracelsus Medical University, Salzburg.
\bibliography{libraryChecked,booksLibrary,ref}
\bibliographystyle{icml2019}

\clearpage
\appendix
This supplementary material contains all proofs omitted in the main submission. 
For readability, all necessary definitions, theorems, lemmas and corollaries are restated (in \textcolor{mydarkblue}{dark blue}) and the numbering matches the original numbering.

Additional (technical) lemmas are prefixed by the section letter, e.g., Lemma~\ref{lem:paring_is_locally_constant}.

\section{Proofs for Section~\ref{section:connectivity_loss}}

First, we recall that the \emph{connectivity loss} is defined as 
\setcounter{equation}{3}
\color{mydarkblue}
\begin{equation}
\label{eqn:connloss}
    \mathcal{L}_{\eta}(S) = \sum\limits_{t \in \deathtimes(S)} |\eta - \varepsilon_t|
\end{equation}
\vspace{-0.1in}

\setcounter{equation}{0}
\color{black}

\setcounter{defi}{2}
\color{mydarkblue}
\begin{defi}
\normalfont
\label{def:ph_indicator_function}
Let $S \subset \mathbb{R}^n$, $|S|=b$ and $z_i \in S$.
We define the indicator function
 \[
 \mathbf{1}_{i, j}(z_1, \dots, z_b) =
 \begin{cases}
 1 & \exists t \in \deathtimes(S): \varepsilon_t = || z_i - z_j ||\\
 0 & \text{else}\enspace,
 \end{cases}
 \]
 where $\{i,j\} \subset [b]$ and $(\varepsilon_k)_{k=1}^M$ is the increasing sequence of all pairwise distance \emph{values} of $S$.
\end{defi}

\color{black}
\setcounter{thm}{0}
\color{mydarkblue}
\begin{thm}
\label{thm:topo_loss_with_ph_indicator_function}
Let $S \subset \mathbb{R}^n$, $|S|=b$, such that the pairwise
distances are unique. Further, let $\mathcal{L}_\eta$ be defined
as in Eq.~\eqref{eqn:top_reg} and $\mathbf{1}_{i,j}$ as
in Definition~\ref{def:ph_indicator_function}. Then,
 \[
 \mathcal{L}_{\eta}(S)
 =
 \sum\limits_{\{i, j\} \subset [b]}
 \big| \eta -  \|z_i - z_j\|\big|\cdot \mathbf{1}_{i,j}(z_1, \dots, z_b)
 \enspace.
 \]
\end{thm}

\color{black}
\vspace{-0.1in}
\begin{proof}
  We have to show that 
  \[
  \sum\limits_{t \in \deathtimes(S)} |\eta - \varepsilon_t|
  \]
  from Eq.~\eqref{eqn:connloss}, denoted as $A$, equals the 
  right-hand side of Theorem~\ref{thm:topo_loss_with_ph_indicator_function},
  denoted as $B$.
  
  \vskip1ex
  \textbf{Part 1} ($A\leq B$). 
  Let $t \in \deathtimes(S)$. 
  Since the pairwise distances of $S$ are unique, $t$ is contained only once in the multi-set $\deathtimes(S)$ and we can treat $\deathtimes(S)$ as an \emph{ordinary} set.  
  Further, there is a (unique) $\{i_t, j_t\}$ such that $\varepsilon_t = \|z_{i_t} - z_{j_t}\|$ and hence $\mathbf{1}_{i_t, j_t}(z_1, \dots, z_b) = 1$. 
  This means every summand in $A$ is also present in $B$.
  As all summands are non-negative, $A \leq B$ follows.
  
  \vskip1ex
  \noindent
  \textbf{Part 2} ($B\leq A$). 
  Consider $\{i, j \} \subset [b]$ contributing to the sum, i.e., $\mathbf{1}_{i, j}(z_1, \dots, z_b) = 1$. 
  By definition 
  \[
    \exists t \in \deathtimes(S): \varepsilon_t = \|z_i - z_j \|
  \]
  and therefore the summand corresponding to $\{i, j\}$ in $B$ is present in $A$. 
  Again, as all summands are non-negative $B \leq A$ follows, which concludes the proof. 
\end{proof}

\begin{lem}
\label{lem:paring_is_locally_constant}
Let $S \subset \R^n$, $|S|=b$, such that the pairwise distances are unique. 
Then, $\mathbf{1}_{i,j}(S)$ is locally constant in $S$. 
Formally, let $1\leq u \leq b, 1\leq v\leq n$, $h \in \R$ and
\[
  S' = \{z_1, \dots, z_{u-1}, z_u + h\cdot e_v, z_{u+1}, \dots, z_b\} 
\]
where $e_v$ is the $v$-th unit vector. Then,  
\[
  \exists \xi > 0 : |h| < \xi \Rightarrow \mathbf{1}_{i,j}(S) = \mathbf{1}_{i,j}(S')
  \enspace. 
\]
\end{lem}
\begin{proof}
$\mathbf{1}_{i,j}(X)$ is defined via $\deathtimes(X)$,  which, in turn, is defined via the Vietoris-Rips filtration of $X$. 
Hence, it is sufficient to show that the corresponding Vietoris-Rips filtrations of $S$ and $S'$ are equal, which we will do next.

Let $(\varepsilon_k)_{k=1}^M$ be the increasing sorted sequence of pairwise distance \emph{values} of $S$. 
As all pairwise distances are unique, there is exactly one $\{i_k, j_k\}$ for each $k$ such that
\[
  \varepsilon_k = \|z_{i_k} - z_{j_k}\|\enspace.
\]
Further, let $S' = \{z'_1, \dots, z'_b\}$ be such that 
\[
  z'_i =
  \begin{cases}
    z_i & i \neq u \\
    z_u + h  \cdot e_v & i = u
  \end{cases}
  \enspace,
\]
and $\varepsilon'_k = \|z'_{i_k} - z'_{j_k}\|$. 
We now show that $(\varepsilon'_k)_{k=1}^M$ is sorted and strictly increasing. 
First, let 
\begin{equation}
  \mu = \min\limits_{1 \leq k < M} \varepsilon_{k+1} - \varepsilon_k
  \enspace. 
\label{eqn:defmu}
\end{equation}
By construction, it follows that $\mu > 0$. 
Now, by the triangle inequality, 
\begin{align*}
  \left|
  \|z'_i - z'_j \| - \|z_i - z_j \| 
  \right|
  & \leq
   \left|
  \|z_i - z_j \|  + |h| - \|z_i - z_j \| 
  \right|\\
  & =
  |h|
  \enspace,
\end{align*}
which is equivalent to 
\[
  -|h| \leq 
  \|z'_i - z'_j \| - \|z_i - z_j \| 
  \leq |h|\enspace.
\]
This yields 
\begin{equation}
\label{eq_1:lem:paring_is_locally_constant}
 \|z'_i - z'_j \| \geq \|z_i - z_j \| - |h|
\end{equation}
and 
\begin{equation}
\label{eq_2:lem:paring_is_locally_constant}
- \|z'_i - z'_j \| \geq - \|z_i - z_j \| - |h|
\enspace.
\end{equation}
Using Eqs. \eqref{eq_1:lem:paring_is_locally_constant} and \eqref{eq_2:lem:paring_is_locally_constant}
we get 
\begin{align*}
\varepsilon'_{k+1} - \varepsilon'_k 
&=
\|z'_{i_{k+1}}-z'_{j_{k+1}}\| - \|z'_{i_{k}}-z'_{j_{k}}\| \\
&\geq 
\|z_{i_{k+1}}-z_{j_{k+1}}\| - |h| - \|z_{i_{k}}-z_{j_{k}}\| - |h|\\
&=
\varepsilon_{k+1} - \varepsilon_{k} - 2 |h|\\ 
&\stackrel{\text{by Eq.}~\eqref{eqn:defmu}}{\geq}
\mu - 2|h| \enspace.
\end{align*}
Overall, $(\varepsilon'_k)_{k=1}^M$ is sorted and strictly increasing if 
\[
  \mu - 2|h| > 0 \Leftrightarrow  |h| < \frac{\mu}{2}\enspace.
\]
It remains to show that the Vietoris-Rips filtration
\[
\emptyset \subset  \mcV_{0}(S)  \subset \mcV_{\nicefrac{\varepsilon_1}{2}}(S) \subset \dots \subset \mcV_{\nicefrac{\varepsilon_M}{2}}(S)
\]
is equal to
\[
\emptyset \subset \mcV_{0}(S') \subset \mcV_{\nicefrac{\varepsilon'_1}{2}}(S') \subset \dots \subset \mcV_{\nicefrac{\varepsilon'_M}{2}}(S')\enspace.
\]
For 
\[
  \mcV_0(S) = \big\{\{1\}, \dots, \{N\}\big\} =\mcV_0(S')
\]
this is obvious. 
For $f_S, f_{S'}$, as in Definition \ref{def:vr_complex} (main paper; Vietoris-Rips complex), we get
\[
f_S^{-1}(\nicefrac{\varepsilon_{k+1}}{2}) = \big\{\{i_k, j_k\}\big\} = f_{S'}^{-1}(\nicefrac{\varepsilon'_{k+1}}{2})
\]
since
\[
  \varepsilon_k = \|z_{i_k} - z_{j_k}\| 
  \text{ and }
  \varepsilon'_k = \|z'_{i_k} - z'_{j_k}\| 
\]
and the pairwise distances are unique.
Now, by induction
\begin{align*}
  \mcV_{\nicefrac{\varepsilon'_{k+1}}{2}}(S') 
  &= 
  \mcV_{\nicefrac{\varepsilon'_{k}}{2}}(S') \cup f_{S'}^{-1}(\nicefrac{\varepsilon'_{k+1}}{2})\\
  &= 
  \mcV_{\nicefrac{\varepsilon_{k}}{2}}(S) \cup f_{S'}^{-1}(\nicefrac{\varepsilon'_{k+1}}{2})\\
  &= 
  \mcV_{\nicefrac{\varepsilon_{k}}{2}}(S) \cup f_{S}^{-1}(\nicefrac{\varepsilon_{k+1}}{2})
  =
  \mcV_{\nicefrac{\varepsilon_{k+1}}{2}}(S) \enspace.
\end{align*}
Setting $\xi = \nicefrac{\mu}{2}$ concludes the proof. 
\end{proof}

\color{mydarkblue}
\begin{thm}
 \label{thm:ph_indicator_loss_function_differentiable}
 Let $S \subset \mathbb{R}^n$, $|S|=b$, such that the pairwise distances are unique.
 Then, for $1\leq u \leq b$ and $1\leq v\leq n$, the partial (sub-)derivative
 of $\mathcal{L}_\eta(S)$ w.r.t. the $v$-th coordinate of $z_u$ exists, i.e.,
 \[
  \frac{\partial \mathcal{L}_{\eta}(S)}{\partial{z_{u, v}}}
  = \sum\limits_{\{i, j\} \subset [b]}
 \frac{\partial \big| \eta -  \|z_i - z_j\|\big|}{\partial{z_{u, v}}} \cdot \mathbf{1}_{i,j}(z_1, \dots, z_b)\enspace.
 \]

\end{thm}

\color{black}
\vspace{-0.1in}
\begin{proof}
  By Theorem \ref{thm:topo_loss_with_ph_indicator_function}, we can write
\[
  \mcL_{\eta}(S)
  =
  \sum\limits_{\{i, j\} \subset [b]}
  \big| \eta -  \|z_i - z_j\|\big|\cdot \mathbf{1}_{i,j}(z_1, \dots, z_b)
  \enspace. 
\]
Further, from Lemma~\ref{lem:paring_is_locally_constant}, we know that $\mathbf{1}_{i,j}$ is locally constant for $u,v$. 
Consequently, the partial derivative w.r.t. $z_{u,v}$ exists and is zero. 
The rest follows from the product rule of differential calculus. 
\end{proof}

\section{Proofs for Section~\ref{section:theory}}

\setcounter{lem}{0}
\color{mydarkblue}
\begin{lem}
  \label{lem:onion_ring}
  Let $2\leq b \leq m$ and $M \subset \mathbb{R}^n$ with $|M| = m$ such that for each $S \subset M$ with $|S| = b$, it holds that $S$ is $\alpha$-$\beta$-connected.
  Then, for $d = m - b$ and $z \in M$ arbitrary but fixed, we find
  $M_z \subset M$ with $|M_z| = d + 1$ and $M_z \subset B(z, \alpha, \beta)$.
\end{lem}

\color{black}
\vspace{-0.1in}
\begin{proof}
  Let $z \in M$. Our strategy is to 
  iteratively construct a set of points 
  \[\{z_1, \dots, z_{d+1} \} \subset B(z, \alpha, \beta) \cap (M\setminus \{z\})\enspace.
  \]
  
  First, consider some $S^{(1)} \subset M$ with $z \in S^{(1)}$ and $|S^{(1)}|=b$.
  Since $S^{(1)}$ is $\alpha$-$\beta$-connected (by assumption), there 
  is $S^{(1)} \ni z_1 \in B(z, \alpha, \beta)$.
   
  By repeatedly considering $S^{(i)} \subset M$ with $z_i \in S^{(i)}$ and 
  $|S^{(i)}|=b$, we
  can construct $M_z^{(i)} = \{z_1, \dots, z_{i}\}$ for $i \leq d = m-b$.
  It holds that 
  \setcounter{equation}{4}
  \begin{equation}
  \label{eq:onion_ring}
  |M\setminus M_z^{(i)}| = m - i \geq m - d = m - (m - b) = b\enspace . 
  \end{equation}
  Hence, we find $S^{(i+1)} \subset M \setminus M_z^{(i+1)}$ with $z \in S^{(i+1)}$ such that $|S^{(i+1)}| = b$. 
  Again, as $S^{(i+1)}$ is $\alpha$-$\beta$-connected, there is
  $S^{(i+1)} \ni z_{i+1} \in B(z, \alpha, \beta)$.
  Overall, this \emph{specific procedure} allows constructing  
  $d+1$ points, as for 
  $i \geq d+1$, Eq. \eqref{eq:onion_ring} is no longer fulfilled. 
\end{proof}

\setcounter{cor}{0}
\color{mydarkblue}
\begin{cor}
\label{cor:onionring_cor2}
  Let $2\leq b \leq m$ and $M \subset \mathbb{R}^n$ with $|M| = m$ such that for each $S \subset M$ with $|S| = b$, it holds that $S$ is $\alpha$-$\beta$-connected. Then
  $M \text{ is } (m - b + 1)\text{-}\beta\text{-dense}$.
\end{cor}

\color{black}
\vspace{-0.1in}
\begin{proof}
  By Lemma~\ref{lem:onion_ring}, we can construct $m-b+1$ points, $M_z$, such that $M_z \subset B(z, \alpha, \beta)$. 
  Conclusively,  
  \[
  y \in M_z \Rightarrow \|z-y\| \leq \beta \enspace.
  \]
\end{proof}

\color{mydarkblue}
\begin{cor}
      \label{cor:onionring_cor1}
  Let $2\leq b \leq m$ and $M \subset \mathbb{R}^n$ with $|M| = m$ such that for each $S \subset M$ with $|S| = b$, it holds that $S$ is $\alpha$-$\beta$-connected.
  Then, for $\varepsilon > 0$ and $m-b+1 > \mathcal{E}^{\varepsilon,n}_{\alpha, \beta}$,
  it follows that $M$ is not $\varepsilon$-separated.
  \end{cor}

\color{black}
\vspace{-0.1in}
\begin{proof}
  Choose some $z \in M$.
  By Lemma~\ref{lem:onion_ring}, we can construct $m-b+1$ points, $M_z$, such that $M_z \subset B(z, \alpha, \beta)$.  
  The distance induced by $\|\cdot\|$ is translation invariant, hence
  \[
  \mathcal{E}^{\varepsilon,n}_{\alpha, \beta} = N_{\varepsilon}\big(B(z, \alpha, \beta)\big)\enspace.
  \] 
  If $m - b + 1 > \mathcal{E}^{\varepsilon,n}_{\alpha, \beta}$, we conclude that $M_z$ is not $\varepsilon$-separated and therefore $M$ is not $\varepsilon$-separated. 
\end{proof}

\color{mydarkblue}
\begin{lem}
\label{lem:onion_ring_entropy}
  Let $\varepsilon < 2\alpha$ and $\alpha < \beta$. Then, in
  $(\mathbb{R}^n, \|\cdot\|_1)$, it holds that
  $\mathcal{E}^{\varepsilon,n}_{\alpha, \beta}
  \leq
  (\nicefrac{2\beta}{\varepsilon}+1)^n -
  (\nicefrac{2\alpha}{\varepsilon}-1)^n$.
\end{lem}

\color{black}
\vspace{-0.1in}
\begin{proof}
  Let $M \subset B(0, \alpha, \beta)$ such that $M$ is $\varepsilon$-separated. 
  Then, the open balls 
  $B^0(z, \nicefrac{\varepsilon}{2})$, $z \in M$, 
  are pairwise disjointly contained in 
  $B(0, \alpha - \nicefrac{\varepsilon}{2}, \beta + \nicefrac{\varepsilon}{2})$. 
  To see this,  let $y \in B^0(z,\nicefrac{\varepsilon}{2})$. 
  We get
  \[
  \|y\| \leq \|y - z \| + \| z \| < \nicefrac{\varepsilon}{2} + \beta
  \]
  and (by the reverse triangle inequality) 
  \begin{align*}
  \| y \| = \|z-(z-y)\| & \geq \big| \|z\| - \|z-y\|  \big| \\
              & \geq \|z\| - \|z-y\| \geq \alpha - \nicefrac{\varepsilon}{2}\enspace.
  \end{align*}
  Hence, $y \in B(0, \alpha, \beta)$. 
  The balls are pairwise disjoint as $M$ is $\varepsilon$-separated and the radius of each ball is chosen as $\nicefrac{\varepsilon}{2}$. 
  Let $\lambda$ denote the Lebesgue measure in $\R^n$. 
  It holds that 
  \begin{align*}
  |M| \cdot \lambda\big(B^0(0, \nicefrac{\varepsilon}{2})\big)
  &= 
  \lambda\left(\bigcup\limits_{z \in M} B^0(z, \nicefrac{\varepsilon}{2})\right)\\
  &\leq \lambda\big(B(0, \alpha - \nicefrac{\varepsilon}{2}, \beta + \nicefrac{\varepsilon}{2})\big)
  \end{align*}
  as $\lambda$ is translation invariant and
  \[
  \bigcup\limits_{z \in M} B^0(z, \nicefrac{\varepsilon}{2}) 
  \subset 
  B(0, \alpha - \nicefrac{\varepsilon}{2}, \beta + \nicefrac{\varepsilon}{2})
  \enspace.
  \]
  The volume of the $\|\cdot\|_1$-ball with radius $r$ is 
  \[
  \lambda\big(B(0, r)\big) 
  = 
  \frac{2^n}{n!}r^n
  \enspace. 
  \]
  Hence, we get 
  \[
  |M| \cdot \frac{\varepsilon^n }{n!}
  \leq
  \frac{2^n}{n!} \left( (\beta+\nicefrac{\varepsilon}{2})^n - (\alpha-\nicefrac{\varepsilon}{2})^n \right) 
  \]
  and thus 
  \begin{align*}
  |M| & \leq \frac{2^n}{\varepsilon^n} \cdot \big((\beta + \nicefrac{\varepsilon}{2})^n - (\alpha-\nicefrac{\varepsilon}{2})^n\big) \\
      & = 
      \frac{2^n}{\varepsilon^n}
      \cdot 
      \frac{\varepsilon^n}{2^n}
      \big(
      (\nicefrac{2\beta}{\varepsilon}+1)^n -  
      (\nicefrac{2\alpha}{\varepsilon}-1)^n
      \big) \\
      & = 
      (\nicefrac{2\beta}{\varepsilon}+1)^n -  
      (\nicefrac{2\alpha}{\varepsilon}-1)^n\enspace.
  \end{align*}
  As the upper bound holds for any $M$, it specifically holds for the largest $M$, which bounds the metric entropy $\mathcal{E}^{\varepsilon,n}_{\alpha, \beta}$ and completes the proof.
  \end{proof}

\section{Parallel persistent homology computation}
\label{section:parallel_ph_computation}

While there exist many libraries for computing persistent homology 
(\texttt{DIPHA} \cite{Bauer14a}, 
\texttt{Dinoysus}\footnote{\scriptsize \url{http://www.mrzv.org/software/dionysus2}}, 
\texttt{JavaPlex}\footnote{\scriptsize \url{https://appliedtopology.github.io/javaplex/}} \cite{Tausz14a}, 
\texttt{GUDHI}\footnote{\scriptsize \url{http://gudhi.gforge.inria.fr}}) 
of a filtered simplicial complex, or fast 
(\texttt{RIPSER}\footnote{\scriptsize \url{https://github.com/Ripser/ripser}}) and approximate (\texttt{SimBa}) \cite{Dey16a} computation of Vietoris-Rips persistent homology, we are not aware of an available  implementation that 

\begin{enumerate}[label=(P\arabic*)]
\item fully operates on the GPU and \label{ph:imp1}
\item offers easy access to the persistence pairings. \label{ph:imp2}
\end{enumerate}

As most deep learning platforms are optimized for GPU computations, \ref{ph:imp1} is important to avoid efficiency bottlenecks caused by expensive data transfer operations between main memory and GPU memory; 
\ref{ph:imp2} is required to enable the integration of persistent homology in an automatic differentiation framework, such as PyTorch. 

Next, we present a straightforward (and not necessarily optimal) variant 
of the \emph{standard} reduction algorithm to compute persistent homology, as introduced in 
\citep[p.~153]{Edelsbrunner2010}, that offers both properties. While many improvements of our parallelization approach are possible, e.g., using \emph{clearing} \cite{Bauer14b} or computing cohomology \cite{deSliva11a} instead, we do not follow these directions here. We only
present a simple parallel variant that is sufficient for the purpose of this work.

The core idea of the original reduction algorithm is to transform the boundary matrix  of a filtered simplicial complex such that the ``birth-death'' times of its homological features can be easily read off.
More precisely, the \emph{boundary matrix} \cite{Edelsbrunner2010} is transformed to its \emph{reduced form} (see Definition \ref{def:low_reduced}) via left-to-right column additions, defined in Algorithm~\ref{alg:column_additions}.

First, we need to define what is meant by a \emph{reduced form} of a boundary matrix $B$ over $\mathbb{Z}_2^{m \times n}$.

\vskip1ex
\begin{defi}
\normalfont
\label{def:low_reduced}
  Let $B \in \Z_2^{m \times n}$ and $B{[i]}$, $B{[\leq i]}$ denote the $i$-th column and the sub-matrix of the first $i$ columns, resp., of $B$.  
  Then, for $B{[j]} \neq 0$, we define
  \[
  \texttt{low}(B, i) = j
  \]
  iff $j$ is the row-index of the lowest $1$ in $B{[i]}$. 
  For convenience, we set 
  \[
  \texttt{low}(B, i) = -1 
  \]
  for $B{[j]} = 0$. 
  We call $B$ \emph{reduced} iff for $1 \leq i < j \leq n$
  \[
  B[i], B[j] \neq 0 \Rightarrow \texttt{low}(B, i) \neq \texttt{low}(B, j) \enspace.
  \]
\end{defi}

\begin{algorithm}
\caption{Column addition \label{alg:column_additions}}
\begin{algorithmic}
  \Function{add}{$B, i, j$}:
    \State $B[j] \leftarrow B[j] + B[i]$ \Comment{Addition in $\mathbb{Z}_2$}
  \EndFunction
\end{algorithmic}
\end{algorithm}

Next, we restate the original (sequential) reduction algorithm.
Let $\partial$ be the boundary matrix of a filtered simplicial complex. 

\begin{algorithm}
\caption{Standard PH algorithm \\ 
\citep[p.~153]{Edelsbrunner2010}\label{alg:standard_ph}}
\begin{algorithmic}
  \State B $\leftarrow \partial$

  \For {$i\leftarrow 1, n$}
  \While{$\exists j_0 < j: \low(B, j_0) = \low(B, j)$}
    \State\Call{add}{$B, j_0, j$}
  \EndWhile
  \EndFor
\end{algorithmic}
\end{algorithm}

Algorithm~\ref{alg:standard_ph} consists of two nested loops. 
We argue that in case column additions would be data-independent, we could easily perform these operations in parallel without conflicts. 
To formalize this idea, let us consider a set $M$ of index pairs
\[
M = \{(i_k, j_k)\}_k  \subset \{1, \dots, b\} \times \{1, \dots, b\}\enspace.
\]

If the conditions
\begin{enumerate}[label=(\roman*)]
\item $\{i_k\}_k \cap \{j_k\}_k = \emptyset$, and
\item $\forall j_k: \exists ! i_k: (i_k, j_k) \in M$
\end{enumerate}
are satisfied, the {\textproc{add}($B, i_k, j_k$) operations from Algorithm~\ref{alg:column_additions} are data-independent. 
Informally, condition (i) ensures that no column is target \emph{and} origin of a merge operation and condition (ii) ensures that each column is targeted by at most one merging operation. 
In the following definition, we construct two auxiliary operators that will allows us to construct $M$ such that conditions (i) and (ii) are satisfied.

\begin{defi}
\normalfont
Let $B \in \Z_2^{m \times n}$ and $1 \leq j \leq m$. 
We define 
\[
I(B, j) = 
\begin{cases}
\emptyset & |\{i: \low(B, i) = j\}| < 2 \\
\{i: \low(i) = j\} & \text{ else }  
\end{cases} 
\]
and
\[
M(B, j) = 
\begin{cases}
\emptyset  &\text{ if }I(B, j) = \emptyset \\
\mu(B,j) \times I(B, j)\setminus \mu(B,j) & \text{ else}
\end{cases}
\]
where $\mu(B,j) = \{\min I(B, j)\}$. Finally, let
\[
M(B) = \bigcup\limits_{j=1}^n M(B, j)
\enspace.
\]
\end{defi}
By construction, it holds that $M(B) = \emptyset$ iff $B$ is reduced.
We can now propose a parallel algorithm, i.e., Algorithm~\ref{alg:gpu_ph}, that iterates until $M(B) = \emptyset$.

\algblock{ParFor}{EndParFor}
\algnewcommand\algorithmicparfor{\textbf{parallel\ for}}
\algnewcommand\algorithmicpardo{\textbf{do}}
\algnewcommand\algorithmicendparfor{\textbf{end\ parallel\ for}}
\algrenewtext{ParFor}[1]{\algorithmicparfor\ #1\ \algorithmicpardo}
\algrenewtext{EndParFor}{\algorithmicendparfor}
\begin{algorithm}
\caption{GPU PH algorithm}
\label{alg:gpu_ph}
\begin{algorithmic}
  \Function{add\ parallel}{$B, M$}:
  \ParFor{$(i, j) \in M$}
    \State\Call{add}{$B, i, j$}
  \EndParFor


  \EndFunction 
\end{algorithmic}
\vskip1ex
\begin{algorithmic}
  \State $B \leftarrow \partial$
  \State $M \leftarrow M(B)$
  \While{$M \neq \emptyset$}
    \State\Call{add\ parallel}{$B, M$}
    \State $M \leftarrow M(B)$
  \EndWhile
\end{algorithmic}
\end{algorithm}

Upon termination, $M(B) = \emptyset$, and hence $B$ is reduced. 
It only remains to show that termination is achieved after a finite number of iterations. 
\begin{lem}
For $B \in \Z_2^{m \times n}$, Algorithm~\ref{alg:gpu_ph} terminates after finitely many iterations. 
\end{lem}
\begin{proof}
Let $B^{(k)}$ be the state of $B$ in the $k$-th iteration. 
For $1 \leq l \leq n$ it holds that $M(B^{(k)}[\leq l]) = \emptyset$ if $B^{(k)}[\leq l]$ is reduced. 
Conclusively, for $k' > k$ 
\[
  B^{(k)}[\leq l] \text{ is reduced } \Rightarrow 
  B^{(k')}[\leq l] \text{ is reduced}
\]
as $B[\leq l]$ does not change any more after the $k$-th iteration.
Hence we can inductively show that the algorithm terminates after finitely many iterations. \\
First, note that $B^{(k)}[\leq 1]$ is reduced. 
Now assume $B^{(k)}[\leq l]$ is reduced and consider
$B^{(k)}[\leq l + 1]$. 
If $B^{(k)}[\leq l + 1]$ is not reduced 
\[
  M\left(B^{(k)}[\leq l + 1]\right) \subset \{1, \dots, l\} \times \{l+1\}
\]
as $B^{(k)}[\leq l]$ is already reduced. 
Thus, if the algorithm continues to the $k+1$-th iteration 
the lowest $1$ of $B^{(k)}[l+1]$ is eliminated and therefore
\[
  \low(B^{(k+1)}, l+1) < \low(B^{(k)}, l+1) \enspace. 
\]
Hence, after $d \leq \low(B^{(k)}, l+1)$ iterations ${B^{(k+d)}[\leq l+1]}$ is reduced as either $B^{(k+d)}[l+1] = 0$ or there is no $j\leq l$ such that 
\[
\low\big(B^{(k+d)}[\leq l], j\big) = \low\big(B^{(k+d)}[\leq l + 1], l+1\big) 
\enspace.
\]
In consequence $B^{(k_0)}[\leq n] = B^{(k_0)}$ is reduced for 
$k_0 < \infty$ which concludes the proof.
\end{proof}

\noindent
\textbf{Runtime study.} We conducted a simple runtime comparison
to \texttt{Ripser} and \texttt{Dionysus} (which both run on the CPU). 
Both implementations are available through
Python wrappers\footnote{\scriptsize For \texttt{Ripser}, see \url{https://scikit-tda.org/}}. \texttt{Dionysus} implements 
persistent cohomology computation \cite{deSliva11a}, while \texttt{Ripser} 
implements multiple recent algorithmic improvements, such as the aforementioned 
clearing optimization as well as computing cohomology. 
Rips complexes are built 
using $\|\cdot \|_1$, up to the 
enclosing radius of the point cloud. 

Specifically, we compute $0$-dimensional features 
on samples of varying size ($b$), drawn from 
a unit multivariate Gaussian in $\mathbb{R}^{10}$. 
Runtime is measured on a 
system with ten Intel(R) Core(TM) i9-7900X CPUs (3.30GHz), 64 GB of RAM and a 
Nvidia GTX 1080 Ti GPU. Figure~\ref{fig:runtime} shows
runtime in seconds, averaged over 50 runs. Note that in this experiment,
runtime includes construction of the Rips complex as well. 


\begin{figure}[t!]
\caption{\label{fig:runtime} Runtime comparison of \texttt{Ripser} \&  
\texttt{Dionysus} (both CPU) vs. our parallel GPU variant. 
Runtime (in seconds) is reported for 
$0$-dimensional VR persistent homology, computed from random samples 
of size $b$ drawn from a unit multivariate Gaussian in $\mathbb{R}^{10}$.}
\vskip0.5ex
\begin{center}
\includegraphics[width=0.99\columnwidth]{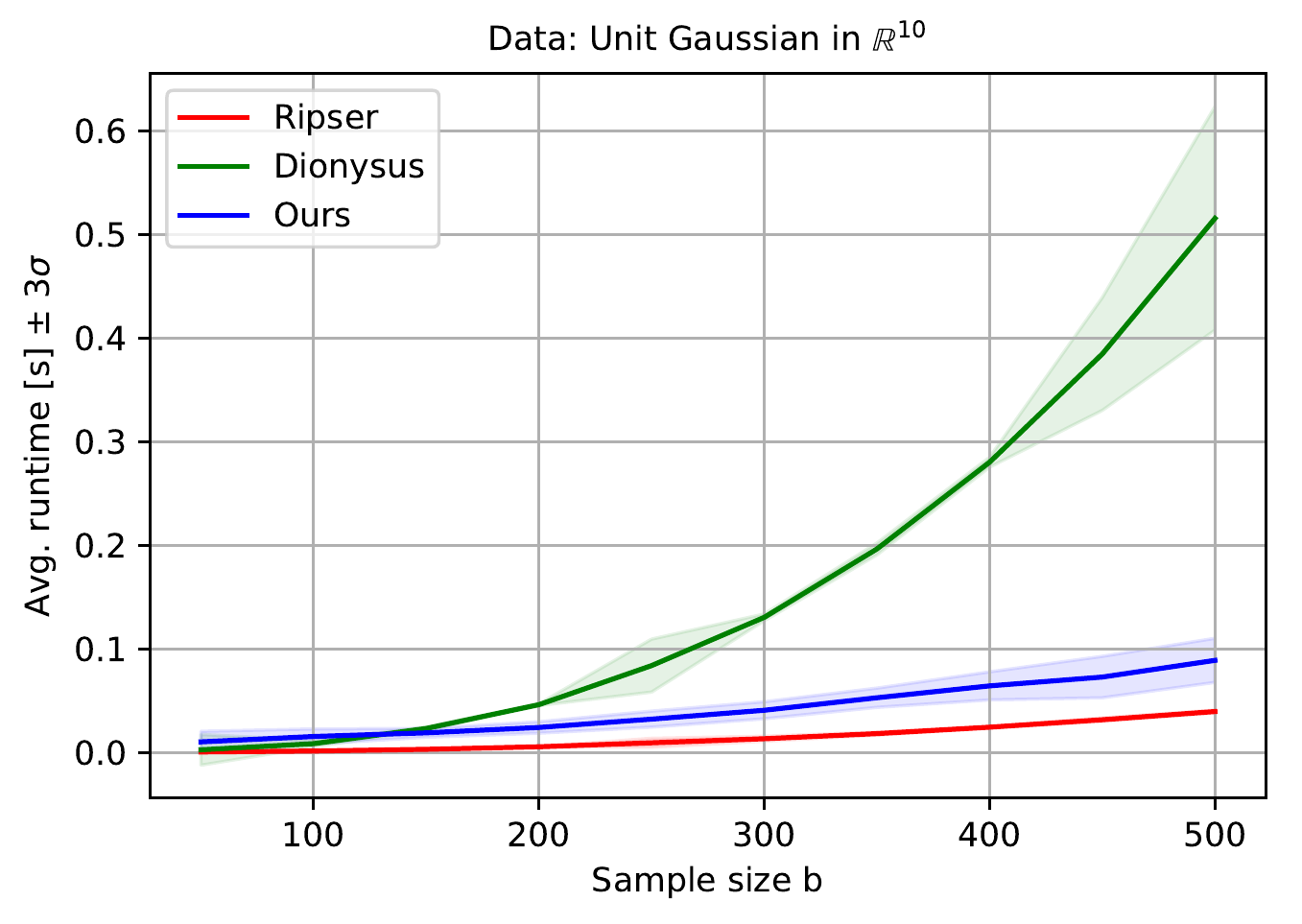}
\end{center}
\vspace{-0.4cm}
\end{figure}

While \texttt{Ripser} is, on average, slightly faster 
than our implementation, we note that for mini-batch sizes 
customary in training neural networks (e.g., 32, 64, 128), the runtime difference is negligible,
especially compared to the overall cost of backpropagation. Importantly, our
method integrates well into existing deep learning frameworks, such
as PyTorch, and thus facilitates to easily experiment with new
loss functions, such as the proposed connectivity loss.

\section{Supplementary figures}

Fig.~\ref{fig:lifetimes_tinyimagenet} shows a second variant of Fig.~6 from the main paper, only that
we replace CIFAR-10 with TinyImage-Net (testing portion). The autoencoder was trained
on the training portion of CIFAR-100.

\begin{figure}
\centering{
\includegraphics[width=0.99\columnwidth]{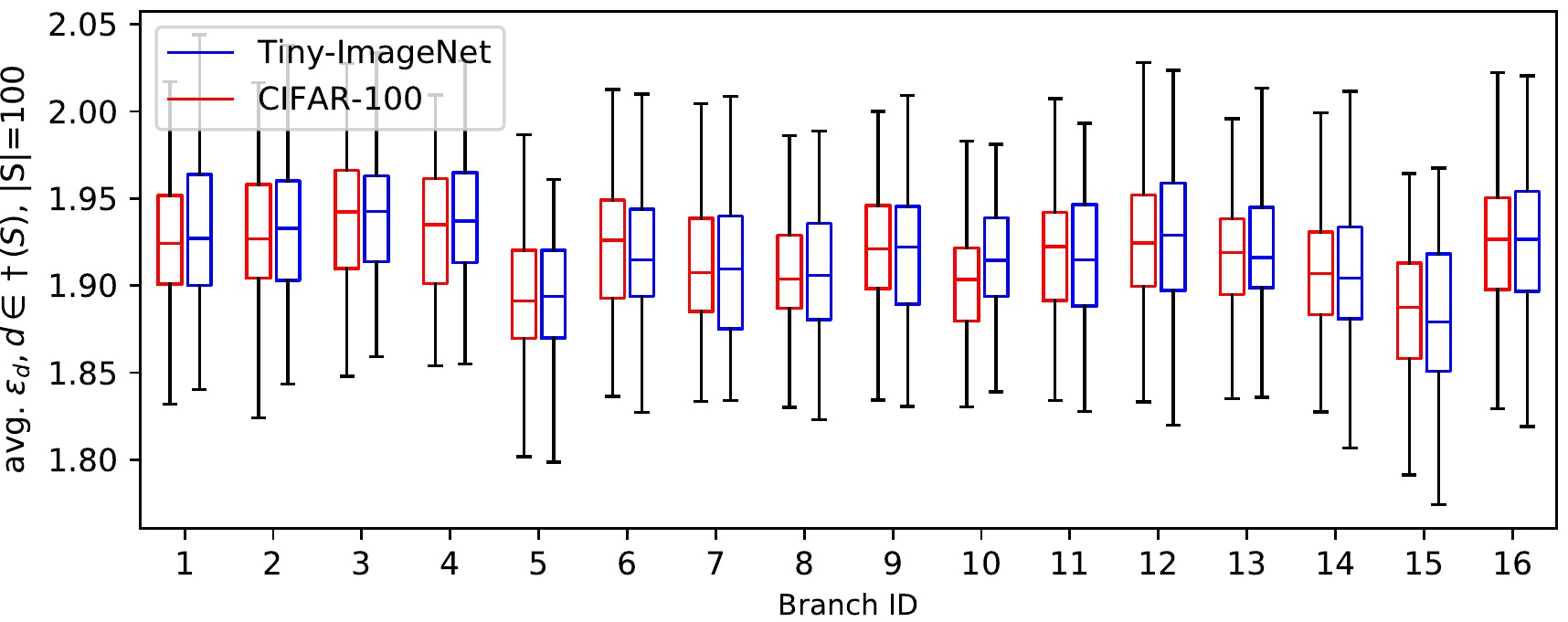}}
\vspace{-0.1in}
\caption{Average $\varepsilon_d, d \in \dagger(S)$, per branch, computed from batches, $S$, of size 100 over
CIFAR-100 (test split) and Tiny-ImageNet (test split); $f_\theta$ is learned from the training portion of CIFAR-100 with $\eta=2$.\label{fig:lifetimes_tinyimagenet}}
\vspace{-0.14in}
\end{figure}

\section{Algorithmic summary}

Algorithm \ref{alg:summary} provides a high-level description of 
the workflow to apply the presented method for one-class learning.

\vskip1ex
\begin{algorithm}[h!]
  \caption{Summary of training steps\label{alg:summary}}
  \textbf{Parameters}: $\eta>0$ (scaling parameter for $\mcL_{\eta}$); 
  $\lambda>0$ (weighting for $\mcL_{\eta}$); $B\geq 1$ (number of branches); 
  $D\geq 1$ (branch dimensionality); $b$ (mini-batch size);

  \vskip1ex
  \noindent
  \textcolor{mydarkblue}{\emph{\underline{Remark}: These are 
  all global parameters.}}
  
\vskip1ex
  \begin{algorithmic}
  \Function{slice}{$z,j$}:
  \State \Return $z[D\cdot(j-1):D \cdot j]$
  \EndFunction 
\end{algorithmic}

\vskip1ex
\textbf{Step 1: Autoencoder training}  
  \begin{algorithmic}
  \State Train $g_{\phi}$ and $f_{\theta}$ using an auxiliary unlabled 
  dataset $\{a_1,\ldots,a_M\}$, minimizing (over batches of size $b$) 
  \[
  \frac{1}{b} \sum_{i=1}^b \|a_i - g_\phi \circ f_\theta(a_i) \|_1 + \textcolor{black}{\lambda
\sum_{j=1}^{B} \mathcal{L}_{\eta}(\{z_1^j,\ldots,z_b^j\})}
  \]
  where $z_i = f_{\theta}(a_i)$ with $z_i^j = \textsc{slice}(z_i,j)$. 
  
  \vskip1ex
  \noindent
  \textcolor{mydarkblue}{\emph{\underline{Remark}: This
  autoencoder can be re-used. That is, if we already have $f_{\theta}$ trained on 
  $\{a_1,\ldots,a_M\}$ (e.g., from another one-class scenario) using the same 
  $\eta, B, D$ parameter choices, autoencoder training can be omitted.}}
  \end{algorithmic} 
 
 \vskip1ex
 \textbf{Step 2: Create one-class model}
 
  \begin{algorithmic}
  \State For one-class samples $\{x_1,\ldots,x_m\}$, compute and store
  $z_i^j = \textsc{slice}(f_{\theta}(x_i),j)$.
  \end{algorithmic}

\vskip1ex
\textbf{Step 3: Evaluate one-class model}
    \begin{algorithmic}
  \State For each new sample $y_*$, obtain $y_*^j = \textsc{slice}(f_{\theta}(y_*),j)$ and
  compute the \emph{one-class score}
  \[
  s(y_*) = \sum_{j=1}^{B} \left|\left\{z_i^j: \|z_*^j - z_i^j\|\leq \eta, 1\leq i\leq m \right\}\right|\enspace.
  \]
  using the stored $z_i^j$ from Step 2.
  \end{algorithmic}
\end{algorithm}

\end{document}